\documentclass[final]{cvpr}

\usepackage{times}
\usepackage{epsfig}
\usepackage{graphicx}
\usepackage{amsmath}
\usepackage{amssymb}
\usepackage[svgnames]{xcolor}
\usepackage{multicol}
\usepackage{multirow}
\usepackage{makecell}
\usepackage{caption}
\usepackage{subcaption}
\usepackage{booktabs}
\usepackage{microtype}
\usepackage{enumitem}
\usepackage{arydshln}
\usepackage[capbesideposition=outside,capbesidesep=quad]{floatrow}		% for captions side-by-side to tables, figures
\usepackage{appendix}
\usepackage[symbol]{footmisc}
\definecolor{darkgreen}{rgb}{0,0.5,0}
\definecolor{frenchblue}{rgb}{0.0, 0.45, 0.73}
\usepackage[pagebackref=true,breaklinks=true,colorlinks,bookmarks=false]{hyperref}
\usepackage[capitalize, nameinlink]{cleveref}

\usepackage{mathtools}
\DeclarePairedDelimiterX{\norm}[1]{\lVert}{\rVert}{#1}
\DeclarePairedDelimiterX{\abs}[1]{\lvert}{\rvert}{#1}
\DeclarePairedDelimiterX{\inner}[1]{\langle }{\rangle}{#1}

\def\cvprPaperID{4862} % *** Enter the CVPR Paper ID here
\def\confYear{CVPR 2021}

\newcommand{\boldparagraph}[1]{\vspace{0.1em}\noindent{\bf #1} }
\newcommand{\bolditem}[1]{\noindent\underline{{\textit{#1}}}}

\DeclareMathOperator*{\argmin}{arg\,min}

\newcommand{\OURS}{SOLD${}^{2}$}

\newcommand{\martin}[1]{} %deactivate showing comments

\begin{document}

%%%%%%%%% TITLE
\title{SOLD${}^{\bf \text{\large{2}}}$: Self-supervised Occlusion-aware Line Description and Detection}

\author{Rémi Pautrat\footnotemark~~${}^1$
\and
\hspace{-0.3cm}
Juan-Ting Lin\footnotemark[\value{footnote}]~~${}^1$
\and
\hspace{-0.3cm}
Viktor Larsson${}^1$
\and
\hspace{-0.3cm}
Martin R. Oswald${}^1$
\and
\hspace{-0.3cm}
Marc Pollefeys${}^{1, 2}$
\and
${}^1$ \normalsize{Department of Computer Science, ETH Zurich}
% \and
% ${}^2$ \normalsize{Department of Electrical Engineering, ETH Zurich}
\and
${}^2$ \normalsize{Microsoft Mixed Reality and AI Zurich lab}
}

\maketitle
\thispagestyle{empty}
\pagestyle{empty}

\footnotetext{* Authors contributed equally.}

%%%%%%%%% ABSTRACT
\begin{abstract}
Compared to feature point detection and description, detecting and matching line segments offer additional challenges.
Yet, line features represent a promising complement to points for multi-view tasks.
Lines are indeed well-defined by the image gradient, frequently appear even in poorly textured areas and offer robust structural cues.
We thus hereby introduce the first joint detection and description of line segments in a single deep network.
Thanks to a self-supervised training, our method does not require any annotated line labels and can therefore generalize to any dataset.
Our detector offers repeatable and accurate localization of line segments in images, departing from the wireframe parsing approach.
Leveraging the recent progresses in descriptor learning, our proposed line descriptor is highly discriminative, while remaining robust to viewpoint changes and occlusions.
We evaluate our approach against previous line detection and description methods on several multi-view datasets created with homographic warps as well as real-world viewpoint changes.
Our full pipeline yields higher repeatability, localization accuracy and matching metrics, and thus represents a first step to bridge the gap with learned feature points methods. Code and trained weights are available at \url{https://github.com/cvg/SOLD2}.
\end{abstract}

%%%%%%%%% BODY TEXT
\section{Introduction}
\label{sec:intro}

\begin{figure}[t]
    \centering
    \includegraphics[width=0.98\columnwidth]{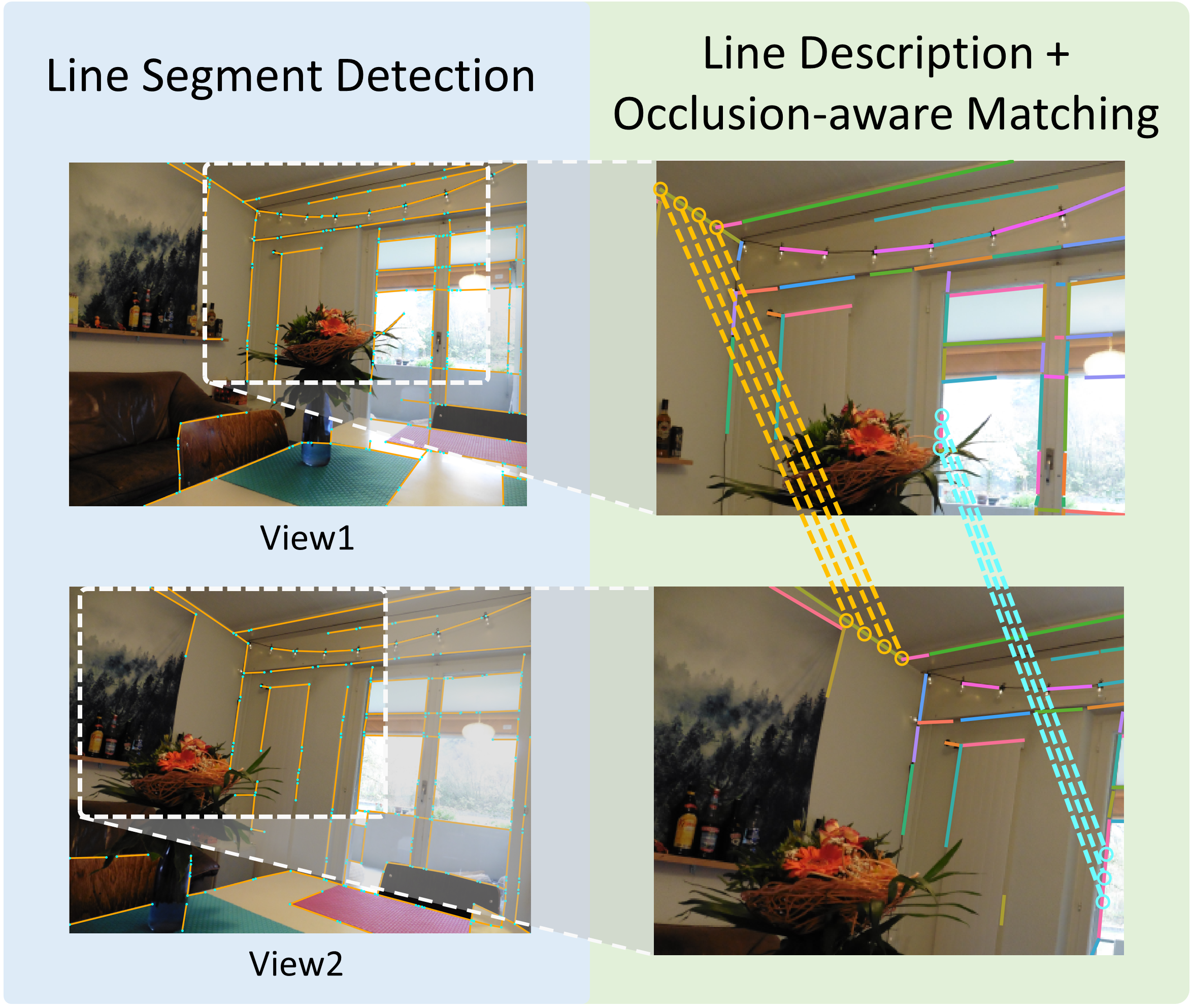}
    \vspace{-4pt}
    \caption{\textbf{Line segment detection and matching.} Our approach detects repeatable lines and is able to match sub-segments to handle partial occlusions. On the right, lines of the same color are matched together.}
    \label{fig:teaser}
\end{figure}

Feature points are at the core of many computer vision tasks such as Structure-from-Motion (SfM)~\cite{heinly2015,schonberger2016structure}, Simultaneous Localization and Mapping (SLAM)~\cite{murartal2015}, large-scale visual localization~\cite{sattler2017,schnberger2017semantic} and 3D reconstruction~\cite{d2net}, due to their compact and robust representation. Yet, the world is composed of higher-level geometric structures which are semantically more meaningful than points. Among these structures, lines can offer many benefits compared to points. Lines are widespread and frequent in the world, especially in man-made environments, and are still present in poorly textured areas. In contrast to points, they have a natural orientation, and a collection of lines provide strong geometric clues about the structure of a scene~\cite{weng1992motion,taylor1995structure,holynski2020}. As such, lines represent good features for 3D geometric tasks.

Previous methods to detect line segments in images often relied on image gradient information and handcrafted filters~\cite{von2008lsd,akinlar2011edlines}. Recently, deep learning has also enabled robust and real-time line detection~\cite{huang2020tp}. Most learned line detectors are however tackling a closely related task: wireframe parsing, which aims at inferring the structured layout of a scene based on line segments and their connectivity~\cite{wireframe,lcnn,hawp,3dwireframe}. These structures provide strong geometric cues, in particular for man-made environments.
Yet, these methods have not been optimized for repeatability across images, a vital feature for multi-view tasks, and their training requires ground truth lines that are cumbersome to manually label~\cite{wireframe}.

The traditional way to match geometric structures across images is to use feature descriptors. Yet, line descriptors face several challenges: line segments can be partially occluded, their endpoints may not be well localized, the scale of the area to describe around each line fluctuates a lot, and it can be severely deformed under perspective and distortion changes~\cite{schmid1997automatic}. Early line descriptors focused on extracting a support region around each line and on computing gradient statistics on it~\cite{wang2009msld,zhang2013lbd}. More recently, motivated by the success of learned point descriptors~\cite{superpoint,d2net,r2d2}, a few deep line descriptors have been proposed~\cite{lange2019dld,vakhitov2019,lange2020wld}. However, they are not designed to handle line occlusion and remain sensitive to poorly localized endpoints.

In this work, we propose to jointly learn the detection and description of line segments. 
To this end, we introduce a self-supervised network, inspired by LCNN~\cite{lcnn} and SuperPoint~\cite{superpoint}, that can be trained on any image dataset without any labels. Pretrained on a synthetic dataset, our method is then generalized to real images. Our line detection aims at maximizing the line repeatability and at being as accurate as possible to allow its use in geometric estimation tasks. The learned descriptor is designed to be robust to occlusions, while remaining as discriminative as the current learned point descriptors. To achieve that, we introduce a novel line matching based on dynamic programming and inspired by sequence alignment in genetics~\cite{needleman1970} and classical stereo matching~\cite{dieny2011}. Thus, our self-supervised occlusion-aware line description and detection (\OURS{}) offers a generic pipeline that aims at bridging the gap with the recent learned feature point methods. Overall, our \textbf{contributions} can be summarized as follows:
\begin{itemize}[topsep=2pt,leftmargin=*,itemsep=0em]
    \item We propose the first deep network for joint line segment detection and description.
    \item We show how to self-supervise our network for line detection, allowing training on any dataset of real images.
    \item Our line matching procedure is robust to occlusion and achieves state-of-the-art results on image matching tasks.
\end{itemize}

\section{Related work}
\label{sec:rw}

\begin{figure*}[ht]
\centering
\includegraphics[width=\textwidth]{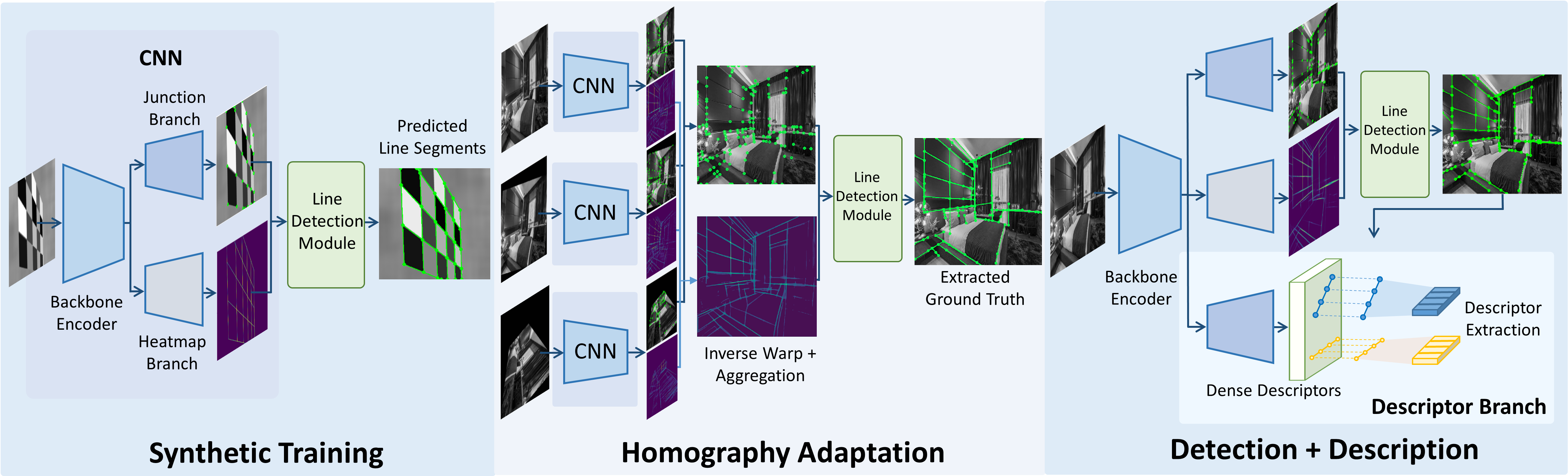}
\vspace{-18pt}
\caption{\textbf{Training pipeline overview.} \textbf{Left:} Our detector network is first trained on a synthetic dataset with known ground truth. \textbf{Middle:} A pseudo ground truth of line segments is then generated on real images through homography adaptation. \textbf{Right:} Finally, the full model with descriptors is trained on real images using the pseudo ground truth.}
\label{fig:system}
\end{figure*}

\boldparagraph{Line detection.}
Gradient-based line segment detection methods such as LSD~\cite{von2008lsd} and EDLines~\cite{akinlar2011edlines} offer a high runtime efficiency, but are not very repeatable under viewpoint and appearance changes. Deep learning is notoriously good at tackling these issues, but learned line detectors have emerged only recently, with the introduction of the wireframe parsing~\cite{wireframe,hawp,afm,almazan2017mcmlsd}. Wireframes are collections of line segments connected by their two endpoints usually labeled by humans~\cite{wireframe}. Wireframes can be parameterized by the line junctions associated with a line verification module~\cite{wireframe,lcnn}, by an attraction field map (AFM)~\cite{afm, hawp}, by a connected graph~\cite{zhang2019ppgnet}, by a root point and two displacements for the endpoints~\cite{huang2020tp} and can benefit from a deep Hough transform prior~\cite{hough1962,deephough}.
Although these methods can extract qualitatively good line segments from images, they have not been trained to produce repeatable lines under viewpoint changes and can still miss some important line landmarks for localization. We take inspiration from them but aim at detecting generic line segments generalizing to most scenes.

\boldparagraph{Line description.}
While early line descriptors are based on simple color histograms~\cite{bay2005wide}, most handcrafted descriptors leverage the image gradient~\cite{wang2009hld, wang2009msld}. The most common approach is thus to extract a line support region around each line and to summarize gradient information in subregions~\cite{wang2009hld, wang2009msld,hirose2012,zhang2013lbd,verhagen2014scale}. Due to its good performance and efficiency, the Line Band Descriptor (LBD) is the most famous of them, but it still suffers from large viewpoint and appearance changes. It is only recently that line description has been tackled with deep learning. One approach is to extract a patch around the line and to compute a low dimensional embedding optimized through a triplet loss, as in DLD~\cite{lange2019dld} and WLD~\cite{lange2020wld}. On the other hand, a line descriptor can be considered as a collection of point descriptors, following the idea of Liu~\etal~\cite{liu2010}. The Learned Line Descriptor (LLD)~\cite{vakhitov2019} thus samples and describes multiple points along each line, and is conceptually the closest previous approach to our method. Designed to be fast and to be used for SLAM, it is however not invariant to rotations and its performance quickly degrades for large viewpoint changes.

\boldparagraph{Joint detection and description of learned features.}
Jointly learned point detectors and descriptors~\cite{lfnet, superpoint, r2d2, aslfeat} propose to share computation between the keypoint detection and description to get fast inference and better feature representations from multi-task learning.
The describe-then-detect trend first computes a dense descriptor map and then extracts the keypoints location from it~\cite{d2net,aslfeat,yang2020ur2kid,yurun2020d2d}. Supervision is provided by either pixel-wise correspondences from SfM~\cite{d2net,aslfeat}, or from image level correspondences only~\cite{yang2020ur2kid}. HF-Net~\cite{hfnet} unifies keypoint detection, local and global description through a multi-task distillation with multiple teacher networks.
Towards the fully unsupervised spectrum, recent methods tightly couple the detector and descriptor learning to output repeatable and reliable points~\cite{unsuperpoint,r2d2,song2020sekd}. On the other hand, Superpoint~\cite{superpoint} first learned the concept of interest points by pretraining a corner detector on a synthetic dataset and later transferring it to real world images.
We adopt here a similar approach extended to line segments.

\boldparagraph{Line matching.}
Beyond simply comparing descriptor similarities, several works tried to leverage higher-level structural cues to guide line matching~\cite{li2016line}.
One approach considers the neighboring lines/points and finds similar patterns across images, for instance through local clusters of lines~\cite{wang2009wide}, intersections between lines~\cite{kim2010novel} or line-junction-line structures~\cite{li2016hierarchical,li2014robust}. However, these methods cannot match isolated lines. Another direction is to find coplanar sets of lines and points and to leverage line-point invariants as well as simple point matching to achieve line matching~\cite{lourakis98,fan2010line,fan2012,ramalingam2015line}. Finally, a last approach consists in matching points sampled along a line, using for example intensity information and epipolar geometry~\cite{schmid1997automatic} or simply point descriptors~\cite{vakhitov2019}. Our work follows this direction but offers a flexible matching of the points along the line, which handles occlusions.
\section{Method}
\label{sec:method}

We propose a unified network to perform line segment detection and description, allowing to match lines across different images. We achieve self-supervision in two steps. Our detector is first pretrained on a synthetic dataset with known ground truth. The full detector and descriptor can then be trained by generating pseudo ground truth line segments on real images using the pretrained model.
We provide an overview of our training pipeline in Figure~\ref{fig:system} and detail its parts in the following sections.

% The problem formulation
\subsection{Problem formulation}
\label{subsec:problem_formulation}
Line segments can be parametrized in many ways: with two endpoints; with a middle point, a direction and a length; with a middle point and offsets for the endpoints; with an attraction field, etc. In this work, we chose the line representation with two endpoints for its simplicity and compatibility with our self-supervision process discussed in Section~\ref{subsec:learning}. For an image $I$ with spatial resolution $h \times w$, we thus consider in the following; the set of all junctions $P=\{p_n\}_{n=1}^{N}$ and line segments $L=\{l_m\}_{m=1}^{M}$ of $I$. A line segment $l_m$ is defined by a pair of endpoints $(e_m^1, e_m^2)\in P^2$.
% \martin{junctions and endpoints are used here synonymously, but actually they are not: An endpoint can be a junction, but not every endpoint is a junction, right?. I also suggest to clearly define for all variables in which spaces they live in.}

% Description of the line junction and heatmaps.
\subsection{Junction and line heatmap inference}
\label{subsec:heatmaps}

Our network takes grayscale images as input, processes them through a shared backbone encoder that is later divided into three different outputs. A junction map $\mathbf{J}$ predicts the probability of each pixel to be a line endpoint, a line heatmap $\mathbf{H}$ provides the probability of a pixel to be on a line, and a descriptor map $\mathbf{D}$ yields a pixel-wise local descriptor. We focus here on the optimization of the first two branches, while the following sections describe their combination to retrieve and match the line segments of an image.

We adopt a similar approach to SuperPoint's keypoint decoder~\cite{superpoint} for the junction branch, where the output is a coarse $\frac{h}{8} \times \frac{w}{8} \times 65$ feature map $\mathbf{J^c}$. Each 65-dimensional vector corresponds to an $8 \times 8$ patch plus an extra ``no junction" dustbin.
We define the ground truth junctions $\mathbf{y} \in \{1, ..., 65\}^{\frac{h}{8} \times \frac{w}{8}}$ indicating the index of the true junction position in each patch. A junction is randomly selected when several ground truth junctions land in the same patch and a value of $65$ means that there is no junction.
The junction loss is then a cross-entropy loss between $\mathbf{J^c}$ and $\mathbf{y}$:
\begin{equation}
    \mathcal{L}_{junc} = \frac{64}{h \times w} \sum_{i,j = 1}^{\frac{h}{8}, \frac{w}{8}}
        -\log\left(\frac{\exp(J^c_{i j y_{ij}})}{\sum_{k=1}^{65} \exp(J^c_{ijk})}\right)
\end{equation}
%
% \martin{the notation is not entirely explanatory, especially with 3 sub-indices.}
At inference time, we perform a softmax on the channel dimension and discard the 65th dimension, before resizing the junction map to get the final $h \times w$ grid.

The second branch outputs a line heatmap $\mathbf{H}$ at the image resolution $h \times w$.
Given a binary ground truth $\mathbf{H^{GT}}$ with a value of 1 for pixels on lines and 0 otherwise, the line heatmap is optimized via a binary cross-entropy loss:
\begin{equation}
    \mathcal{L}_{line} = \frac{1}{h \times w} \sum_{i,j = 1}^{h, w} - H^{GT}_{ij} \log(H_{ij})
\end{equation}
%
% Line detection method
%------------------------------------------------------------------------------
\subsection{Line detection module}  \label{subsec:line_detection}
%------------------------------------------------------------------------------
%
After inferring the junction map $\mathbf{J}$ and line heatmap $\mathbf{H}$, we threshold $\mathbf{J}$ to keep the maximal detections and apply a non-maximum suppression (NMS) to extract the segment junctions $\hat{P}$.
The line segment candidates set $\hat{L}_{cand}$ is composed of every pair of junctions in $\hat{P}$.
Extracting the final line segment predictions $\hat{L}$ based on $\mathbf{H}$ and $\hat{L}_{cand}$ is non-trivial as the activations along a segment defined by two endpoints may vary a lot across different candidates.
Our approach can be broken down into four parts: (1) regular sampling between endpoints, (2) adaptive local-maximum search, (3) average score, and (4) inlier ratio.
% Extracting the final line segment predictions $\hat{L}$ from $\mathbf{H}$ and $\hat{L}_{cand}$ is non-trivial as the activations along a segment defined by two endpoints may vary a lot across different candidates.
% Instead of adopting a single threshold on the pixels after rasterization~\cite{3dwireframe}, our method can be broken down into four parts: (1) regular sampling between endpoints, (2) adaptive local-maximum search, (3) average score, and (4) inlier ratio.

\bolditem{Regular sampling between endpoints:} Instead of fetching all the rasterized pixels between the two endpoints~\cite{3dwireframe}, we sample $N_s$ uniformly spaced points (including the two endpoints) along the line segment.

\bolditem{Adaptive local-maximum search:} Using bilinear interpolation to fetch the heatmap values at the extracted points $q_k$ may discard some candidates due to the misalignment between the endpoints and the heatmap, especially for long lines. 
To alleviate that, we search for the local maximal heatmap activation $h_k$ around each sampled location $q_k$ within a radius $r$ proportional to the length of the line.

\bolditem{Average score:} The average score is defined as the mean of all the sampled heatmap values: $y_{avg} = \frac{1}{N_s} \sum_{k=1}^{N_s} h_k$. Given a threshold $\xi_{avg}$, valid line segment candidates should satisfy $y_{avg} \geq \xi_{avg}$.

\bolditem{Inlier ratio:} Only relying on the average score may keep segments with a few high activations but with holes along the line. To remove these spurious detections, we also consider an inlier ratio $y_{inlier} = \frac{1}{N_s} \abs{\{h_k | h_k \geq \xi_{avg}, h_k \in H\}}$. Given an inlier ratio threshold $\xi_{inlier}$, we only keep candidates satisfying $y_{inlier} \geq \xi_{inlier}$.

% Self-supervised learning
%------------------------------------------------------------------------------
\subsection{Self-supervised learning pipeline}  \label{subsec:learning}
%------------------------------------------------------------------------------
%
Inspired by the success of DeTone~\etal~\cite{superpoint}, we extend their homography adaptation to the case of line segments.
% For keypoints, one can aggregate probability maps and use NMS to extract reliable interest points. However, line segments are defined by a collection of two endpoints, so one cannot directly perform the same pixel-wise aggregation as for points. Defining a line distance metric and clustering close line segments together would lose the efficiency of the aggregation on a grid of pixels. We instead first aggregate the junction and line heatmaps as in SuperPoint using a set of $N_h$ homographies, and later recombine them to get the refined line segments.
Let $f_{junc}$ and $f_{heat}$ represent the forward pass of our network to compute the junction map and the line heatmap. We start by aggregating the junction and heatmap predictions as in SuperPoint using a set of $N_h$ homographies $(\mathcal{H}_i)_{i=1}^{N_h}$:
\begin{align}
    \mathbf{\hat{J}}(I;f_{junc}) &= \frac{1}{N_h} \sum_{i=1}^{N_h} \mathcal{H}^{-1}_i\big(f_{junc}(\mathcal{H}_i(I))\big)\\
    \mathbf{\hat{H}}(I;f_{heat}) &= \frac{1}{N_h} \sum_{i=1}^{N_h} \mathcal{H}^{-1}_i\big(f_{heat}(\mathcal{H}_i(I))\big)
\end{align}
We then apply the line detection module to the aggregated maps $\mathbf{\hat{J}}$ and $\mathbf{\hat{H}}$ to obtain the predicted line segments $\hat{L}$, which are then used as ground truth for the next training round. Figure~\ref{fig:system} provides an overview of the pipeline. Similar to Superpoint, this process can be iteratively applied to improve the label quality. However, we found that a single round of adaptation already provides sufficiently good labels.

% Line description
%------------------------------------------------------------------------------
\subsection{Line description}  \label{subsec:line_description}
%------------------------------------------------------------------------------
%
Describing lines in images is a problem inherently more difficult than describing feature points. A line can be partially occluded, its endpoints are not always repeatable across views, and the appearance of a line can significantly differ under viewpoint changes. To tackle these challenges, we depart from the classical description of a patch centered on the line~\cite{lange2019dld,lange2020wld}, that is not robust to occlusions and endpoints shortening. Motivated by the success of learned point descriptors, we formulate our line descriptor as a sequence of point descriptors sampled along the line. Given a good coverage of the points along the line, even if part of the line is occluded, the points on the non-occluded part will store enough line details and can still be matched.

The descriptor head of our network outputs a descriptor map $\mathbf{D} \in \mathbb{R}^{\frac{h}{4} \times \frac{w}{4} \times 128}$ and is optimized through the classical point-based triplet loss~\cite{balntas2016,mishchuk2017} used in other dense descriptors~\cite{d2net}. Given a pair of images $I_1$ and $I_2$ and matching lines in both images, we regularly sample points along each line and extract the corresponding descriptors $(\mathbf{D_1^i})_{i=1}^n$ and $(\mathbf{D_2^i})_{i=1}^n$ from the descriptor maps, where $n$ is total number of points in an image. The triplet loss minimizes the descriptor distance of matching points and maximizes the one of non-matching points. The positive distance is defined as
\begin{equation}
    p_i = ||\mathbf{D_1^i} - \mathbf{D_2^i}||_2 
\end{equation}
The negative distance is computed between a point and its hardest negative example in batch:
\begin{equation}
    n_i = \min \Big(||\mathbf{D_1^i} - \mathbf{D_2^{h_2(i)}}||_2, \; ||\mathbf{D_1^{h_1(i)}} - \mathbf{D_2^i}||_2\Big)
\end{equation}
where $h_1(i) = \argmin_{\scriptsize{k \in [1, n]}} ||\mathbf{D_1^k} - \mathbf{D_2^i}||$ such that the points $i$ and $k$ are at a distance of at least $T$ pixels and are not part of the same line, and similarly for $h_2(i)$. The triplet loss with margin $M$ is then defined as
\begin{equation}
    \mathcal{L}_{desc} = \frac{1}{n} \sum_{i=1}^n \max(0, M + p_i - n_i)
\end{equation}

% Total loss
%------------------------------------------------------------------------------
\subsection{Multi-task learning}  \label{subsec:multi-task}
%------------------------------------------------------------------------------
%
Detecting and describing lines are independent tasks with different homoscedastic aleatoric uncertainties and their respective losses can have different orders of magnitude.
Thus, we adopt the multi-task learning proposed by Kendall~\etal~\cite{uncertainty_dynamic} with a dynamic weighting of the losses, where the weights $w_{junc}$, $w_{line}$ and $w_{desc}$ are optimized during training~\cite{Kendall2017,hfnet}. The total loss becomes:
\begin{equation}
    \begin{split}
     \mathcal{L}_{total} =\ &e^{-w_{junc}} \mathcal{L}_{junc} + e^{-w_{line}} \mathcal{L}_{line} \\
                 &+ e^{-w_{desc}} \mathcal{L}_{desc} + w_{junc} + w_{line} + w_{desc}
    \end{split}
\end{equation}

% Line matching
\subsection{Line matching}
\label{subsec:line_matching}
At inference time, two line segments are compared based on their respective collection of point descriptors sampled along each line. However, some of the points might be occluded or, due to perspective changes, the length of a line can vary and the sampled points may be misaligned. The ordering of the points matched along the line should nevertheless be constant, i.e. the line descriptor is an ordered sequence of descriptors, not just a set. To solve this sequence assignment problem, we take inspiration from nucleotide alignment in bioinformatics~\cite{needleman1970} and pixel alignment along scanlines in stereo vision~\cite{dieny2011}. We thus propose to find the optimal point assignment through the dynamic programming algorithm originally introduced by Needleman and Wunsch~\cite{needleman1970}.
% We thus propose to find the optimal point assignment along a line through the dynamic programming algorithm originally introduced by Needleman and Wunsch~\cite{needleman1970} with a few modifications.

\begin{figure}
    \centering
    \includegraphics[width=\columnwidth]{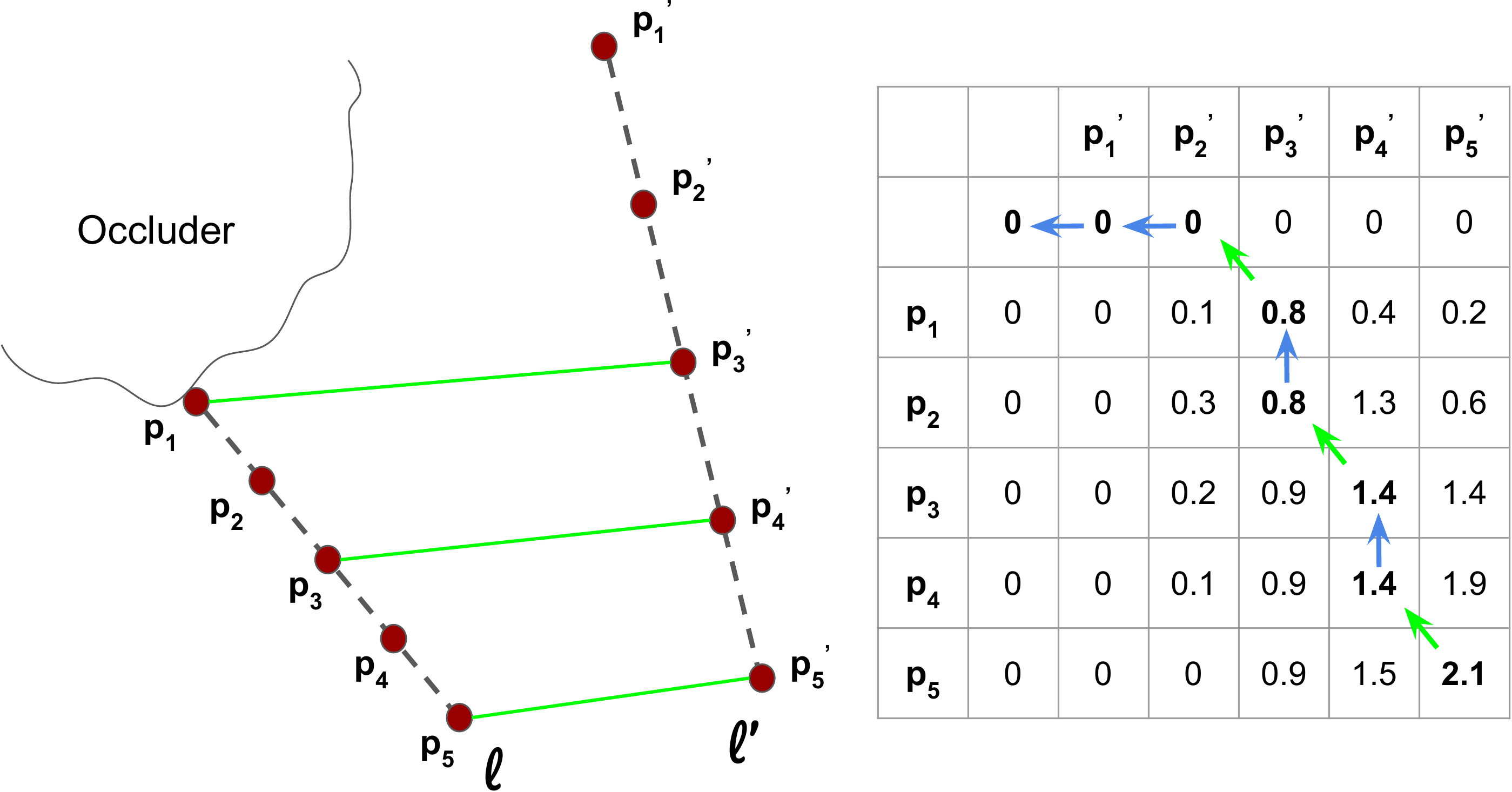}
    \vspace{-22pt}
    \caption{\textbf{Computation of a line match score.} The optimal path selected by the Needleman-Wunsch algorithm is shown in green for matches and blue for skipping a point, using here a $gap$ score of zero.}
    \label{fig:needleman_wunsch_matching}
\end{figure}

When matching two sequences of points, each point can be either matched to another one or skipped. The score attributed to a match of two points depends on the similarity of their descriptors (i.e. their dot product), so that a higher similarity gives a higher score. Skipping a point is penalized by a $gap$ score, which has to be adjusted so that it is preferable to match points with high similarity but to skip the ones with low similarity. The total score of a line match is then the sum of all skip and match operations of the line points. The Needleman-Wunsch (NW) algorithm returns the optimal matching sequence maximizing this total score. This is achieved with dynamic programming by filling a matrix of scores row by row, as depicted in Figure~\ref{fig:needleman_wunsch_matching}. Given a sequence of $m$ points along a line $l$, $m'$ points along $l'$, and the associated descriptors $\mathbf{D}$ and $\mathbf{D'}$, this score matrix $\mathbf{S}$ is an $(m+1) \times (m'+1)$ grid where $\mathbf{S}(i, j)$ contains the optimal score for matching the first $i$ points of $l$ with the first $j$ points of $l'$. The grid is initialized by the $gap$ score in the first row and column, and is sequentially filled row by row, using the scores stored in the left, top and top-left cells:
\begin{equation}
\begin{split}
    \mathbf{S}(i, j) = \max\big(& \mathbf{S}(i-1, j) + gap,\mathbf{S}(i, j-1) + gap,\\
                            & \mathbf{S}(i-1, j-1) + \mathbf{D^i}^T \mathbf{D'^j}\big)
\end{split}
\end{equation}

Once the matrix is filled, we select the highest score in the grid and use it as a match score for the candidate pair of lines. Each line of the first image is then matched to the line in the second image with the maximum match score.
% Note that starting from the highest scoring cell of the grid, one can backtrack in the grid to determine which points were actually matched. The NW algorithm is thus actually able to match sub-segments of the full line, which is essential when a line is partially occluded.

% Implementation details
%------------------------------------------------------------------------------
\subsection{Implementation details}  \label{subsec:implementation}
%------------------------------------------------------------------------------
%
\boldparagraph{Network implementation.}
To have a fair comparison with most wireframe parsing methods~\cite{lcnn, hawp, deephough}, we use the same stacked hourglass network~\cite{hourglass} for our backbone. The three branches of our network are then series of convolutions, ReLU activations and upsampling blocks via subpixel shuffles~\cite{subpixel}. Please refer to the supplementary material for more details about the architecture. The network is optimized with the Adam solver~\cite{kingma2014} with a learning rate of $0.0005$.

\boldparagraph{Line parameters.}
We use a junction threshold of $\frac{1}{65}$, a heatmap threshold $\xi_{avg} = 0.25$, an inlier threshold $\xi_{inlier} = 0.75$, we extract $N_s = 64$ samples along each line to compute the heatmap and inlier scores, and we use $N_h = 100$ homographies for the homography adaptation.

\boldparagraph{Matching details.}
The line descriptor is computed by regularly sampling up to 5 points along each line segment, but keeping a minimum distance of 8 pixels between each point. Since the ordering of the points might be reversed from one image to the other, we run the matching twice with one point-set flipped. A $gap$ score of $0.1$ empirically yields the best results during the NW matching. To speed up the line matching, we pre-filter the set of line candidates with a simple heuristic. Given the descriptor of the 5 points sampled on a line of $I_1$ to be matched, we compute the similarity with their nearest neighbor in each line of $I_2$, and average these scores for each line. This yields a rough estimate of the line match score, and we keep the top 10 best lines as candidates for the NW matching. Finally, we retain at matching time only the pairs that are mutually matched.

\boldparagraph{Training dataset.}
% Our synthetic dataset consists of rendered 2D shapes including triangles, lines, rectangles, and stripes, similarly as in Superpoint~\cite{superpoint}.
% We label the corners of these shapes as junctions and edges as line segments. Gaussian noise, salt and pepper noise, and random shades are added to the rendered images as data augmentation. 
% We generate $20,000$ training samples and $400$ testing samples.
We use the same synthetic dataset as in SuperPoint~\cite{superpoint}, labelling the corners of the geometrical shapes as junctions and edges as line segments.
For the training with real images, we use the Wireframe dataset~\cite{wireframe}, allowing a fair comparison with the current state of the art also trained on these images. 
% This dataset contains $5,462$ images of indoor and outdoor man-made environments. 
We follow the split policy in LCNN~\cite{lcnn}: $5,000$ images for training and $462$ images for testing.
We however only use the images and ignore the ground truth lines provided by the dataset.
\section{Experiments}
\label{sec:exp}

\begin{table*}[ht]
    \centering
    %\footnotesize
    %\setlength{\tabcolsep}{5.2pt}
    \scriptsize
    \setlength{\tabcolsep}{6.4pt}
    \renewcommand{\arraystretch}{1}
    \begin{tabular}{lcccccccccccc}
    \toprule
     & \multicolumn{6}{c}{Wireframe Dataset~\cite{wireframe}} & \multicolumn{6}{c}{YorkUrban Dataset~\cite{yorkurban}} \\ %\cline{2-13}
    \cmidrule(lr){2-7} \cmidrule(lr){8-13}
                  & \multicolumn{2}{c}{$\operatorname{d_{s}}$}  & \multicolumn{2}{c}{$\operatorname{d_{orth}}$} & \multirow{2}{*}{time$\downarrow$} &\multirow{2}{*}{\makecell{\# lines\\/ image}} &
                  \multicolumn{2}{c}{$\operatorname{d_{s}}$}  & \multicolumn{2}{c}{$\operatorname{d_{orth}}$} & \multirow{2}{*}{time$\downarrow$} &\multirow{2}{*}{\makecell{\# lines\\/ image}} \\ %\cline{2-5} \cline{8-11}
                  \cmidrule(lr){2-3} \cmidrule(lr){4-5} \cmidrule(lr){8-9} \cmidrule(lr){10-11}
                  & $\mathrm{Rep\text{-}}5$ $\uparrow$ & $\mathrm{LE\text{-}5}$ $\downarrow$ & $\mathrm{Rep\text{-}5}$ $\uparrow$ & $\mathrm{LE\text{-}5}$ $\downarrow$ & & &
                  $\mathrm{Rep\text{-}5}$ $\uparrow$ & $\mathrm{LE\text{-}5}$ $\downarrow$ & $\mathrm{Rep\text{-}5}$ $\uparrow$ & $\mathrm{LE\text{-}5}$ $\downarrow$ & & \\
     \midrule
     LCNN~\cite{lcnn} @0.98             & 0.434 & 2.589      & 0.570  & 1.725 & 0.120 & 76             & 0.318 & 2.662  & 0.449 & 1.784 & 0.206 & 103  \\
     HAWP~\cite{hawp} @0.97             & 0.451 & 2.625      & 0.537  & 1.738 & 0.035 & 47            & 0.295 & 2.566  & 0.368 & 1.757 & 0.045 & 59   \\
     DeepHough~\cite{deephough} @0.9    & 0.419 & 2.576      & 0.618 & 1.720 & 0.289 & 135           & 0.315 & 2.695  & 0.535 & 1.751 & 0.519 & 206  \\
     TP-LSD~\cite{huang2020tp} HG       & 0.358 & 3.220      & 0.647  & 2.212 & 0.038 & 72            & 0.233 & 3.357  & 0.524 & 2.395 & 0.038 & 113  \\
     TP-LSD~\cite{huang2020tp} TP512    & 0.563 & 2.467      & 0.746  & 1.450 & 0.097 & 81            & 0.433 & 2.612  & 0.633 & 1.555 & 0.099 & 125  \\
     LSD~\cite{von2008lsd}              & 0.358 & 2.079      & 0.707  & 0.825 & \textbf{0.026} & 228	& 0.357 & 2.116  & 0.704 & 0.876 & \textbf{0.031} & 359  \\[0.8pt] \hdashline \noalign{\vskip 1pt}
    % 
    %  Ours w/ NMS  & 0.562	& 0.636  & 1.87  & 2.58          & 0.818  & 0.829  & 1.034	& 1.114     & 0.480 & 126.2 \\
     Ours w/ CS                         & 0.557 & \textbf{1.995}      & 0.801  & 1.119 & 0.042 & 116             & 0.528 & \textbf{1.902}  & 0.787 & 1.107 & 0.064 & 222\\
     Ours                               & \textbf{0.616}	& 2.019 & \textbf{0.914} & \textbf{0.816} & 0.074	& 447           & \textbf{0.582} & 1.932 &\textbf{0.913} &\textbf{0.713} & 0.093 & 1085 \\
     \bottomrule
     \end{tabular}
     \caption{\textbf{Line detection evaluation on the Wireframe~\cite{wireframe} and YorkUrban~\cite{yorkurban} datasets.} We compare repeatability and localization error for an error threshold of 5 pixels in structural and orthogonal distances. Our approach provides the most repeatable and accurate line detections compared to the other baselines.}
     \label{tab:detection_eval}
\end{table*}

% Line segment detection evaluation
%------------------------------------------------------------------------------
\subsection{Line segment detection evaluation}  \label{subsec:line_detection_eval}
%------------------------------------------------------------------------------
%
To evaluate our line segment detection, we use the test split of the Wireframe dataset~\cite{wireframe} and the YorkUrban dataset~\cite{yorkurban}, which contains 102 outdoor images. For both datasets, we generate a fixed set of random homographies and warp each image to get a pair of matching images.

% Line segment distance measure
\boldparagraph{Line segment distance metrics.}
A line distance metric needs to be defined to evaluate the accuracy of a line detection. We use the two following metrics:

\bolditem{Structural distance ($\operatorname{\mathbf{d_s}}$):} The structural distance of two line segments $l_1$ and $l_2$ is defined as:
\begin{equation}
    \begin{split}
        \operatorname{d_s}(l_1, l_2) = \min(
        &\norm{e_1^1 - e_2^1}_2 + \norm{e_1^2 - e_2^2}_2, \\
        &\norm{e_1^1 - e_2^2}_2 + \norm{e_1^2 - e_2^1}_2)
    \end{split}
\end{equation}
where $(e_1^1, e_1^2)$ and $(e_2^1, e_2^2)$ are the endpoints of $l_1$ and $l_2$ respectively. Contrary to the formulation of recent wireframe parsing works~\cite{lcnn, hawp}, we do not use square norms to make it directly interpretable in terms of endpoints distance.
% This definition  measuring the endpoints distance is adopted in recent wireframe parsing works~\cite{lcnn, hawp} to compute the structural AP (sAP).

\bolditem{Orthogonal distance ($\operatorname{\mathbf{d_{orth}}}$):} The orthogonal distance of two line segments $l_1$ and $l_2$ is defined as the average of two asymmetrical distances $d_a$:
\begin{equation}
    \operatorname{d_a}(l_i, l_j) = \norm{e_j^1 - p_{l_i}(e_j^1)}_2
                                   + \norm{e_j^2 - p_{l_i}(e_j^2)}_2
\end{equation}
\begin{equation}
    \operatorname{d_{orth}}(l_1, l_2) = \frac{\operatorname{d_a}(l_1, l_2) + \operatorname{d_a}(l_2, l_1)}{2}
\end{equation}
where $p_{l_j}(.)$ denotes the orthogonal projection on line $l_j$.
% To exclude cases with $l_1$ and $l_2$ having a small orthogonal distance but small overlap region, we also verify that the overlap ratio of the line segments is greater or equal than a threshold $\zeta = 0.5$. The overlap is computed symmetrically and averaged between the two lines.
When searching the nearest line segment with this distance, we ignore the line segments with an overlap below $0.5$.
This definition allows line segments corresponding to the same 3D line but with different line lengths to be considered as close, which can be useful in localization tasks~\cite{micusik2014relaxed}.
% This distance metric is thus useful for 3D reconstruction and localization tasks that mainly rely on such 3D lines.

% Line segment detection metrics
\boldparagraph{Line segment detection metrics.}
Since the main objective of our line segment detection method is to extract repeatable and reliable line segments from images, evaluating it on the manually labeled lines of the wireframe dataset~\cite{wireframe} is not suitable. 
We thus instead adapt the detector metrics proposed for SuperPoint~\cite{superpoint} to line segments using pairs of images.

\bolditem{Repeatability:} The repeatability measures how often a line can be re-detected in different views. It is the average percentage of lines in the first image that have a matching line when reprojected in the second image. Two lines are considered to be matched when their distance is lower than a threshold $\epsilon$. This metric is computed symmetrically across the two images and averaged.

\bolditem{Localization error:} The localization error with tolerance $\epsilon$ is the average line distance between a line and its re-detection in the second image, only considering the matched lines.

% Line segment detection evaluation
\boldparagraph{Evaluation on the Wireframe and YorkUrban datasets.}
We compare in Table~\ref{tab:detection_eval} our line segment detection method with 5 baselines including the handcrafted Line Segment Detection (LSD)~\cite{von2008lsd}, wireframe parsing methods such as LCNN~\cite{lcnn}, HAWP~\cite{hawp}, TP-LSD~\cite{huang2020tp}, and Deep Hough-transform Line Priors (DeepHough)~\cite{deephough}.
LSD is used with a minimum segment length of 15 pixels. For LCNN, HAWP, and DeepHough, we chose thresholds (0.98, 0.97, and 0.9 respectively) on the line scores to maximize their performances. We show two TP-LSD variants: HG using the same backbone~\cite{hourglass} as the other wireframe parsing baselines and our method, and TP512 that uses a ResNet34~\cite{resnet} backbone. 

Overall, our method achieves the best performance in terms of repeatability and localization error on both datasets. We also include our method with candidate selection (CS), which removes the segments having other junctions between the two endpoints to avoid overlapping segments in the predictions $\hat{L}$. Without overlapping segments, the performance slightly decreases but we get fewer segments and faster inference speed. The candidate selection is also used in our descriptor evaluation section and is referred as line NMS.

% Line segment description evaluation
%------------------------------------------------------------------------------
\subsection{Line segment description evaluation}  \label{subsec:line_description_eval}
%------------------------------------------------------------------------------
%
% Line segment descriptor metrics
\boldparagraph{Line descriptor metrics.}
Our line descriptor is evaluated on several matching metrics, both on hand-labeled line segments and on detected line segments (LSD or our predicted lines). 
When using ground truth lines, there is an exact one-to-one line correspondence. 
For predicted lines, ground truth matches are computed with a threshold $\epsilon$ similarly as for the detector metrics. 
When depth is available, the lines are projected to 3D and directly compared in 3D space.
Only lines with a valid reprojection in the other image are considered.

\bolditem{Accuracy:} Percentage of correctly matched lines given a set of ground truth line matches.

\bolditem{Receiver operating characteristic (ROC) curve:} Given a set of matching lines, we compute the SIFT~\cite{lowe2004sift} descriptor of each endpoint, average the SIFT distances between each pair of lines, and use the second nearest neighboring line as negative match. The ROC curve is then the true positive rate (TPR) plotted against the false positive rate (FPR). The curve is obtained by varying the descriptor similarity threshold defining a positive match.

\bolditem{Precision:} Ratio of true positive matches over the total number of predicted matches.

\bolditem{Recall:} Ratio of true positive matches over the total number of ground truth matches.

\begin{figure}
    \begin{minipage}{0.48\columnwidth}
        \footnotesize
        \centering
        \setlength{\tabcolsep}{4.0pt}
        \renewcommand{\arraystretch}{1.4}
        \begin{tabular}{lc}
            \toprule
            Method & Accuracy$\uparrow$ \\
            \midrule
            LBD~\cite{zhang2013lbd} & 0.610 \\
            LLD~\cite{vakhitov2019} & 0.265 \\
            WLD~\cite{lange2020wld} & 0.933 \\
            \OURS{} (Ours) & \textbf{0.978} \\
            \bottomrule
        \end{tabular}
    \end{minipage}
    \begin{minipage}{0.51\columnwidth}
        \centering
        \includegraphics[width=\columnwidth]{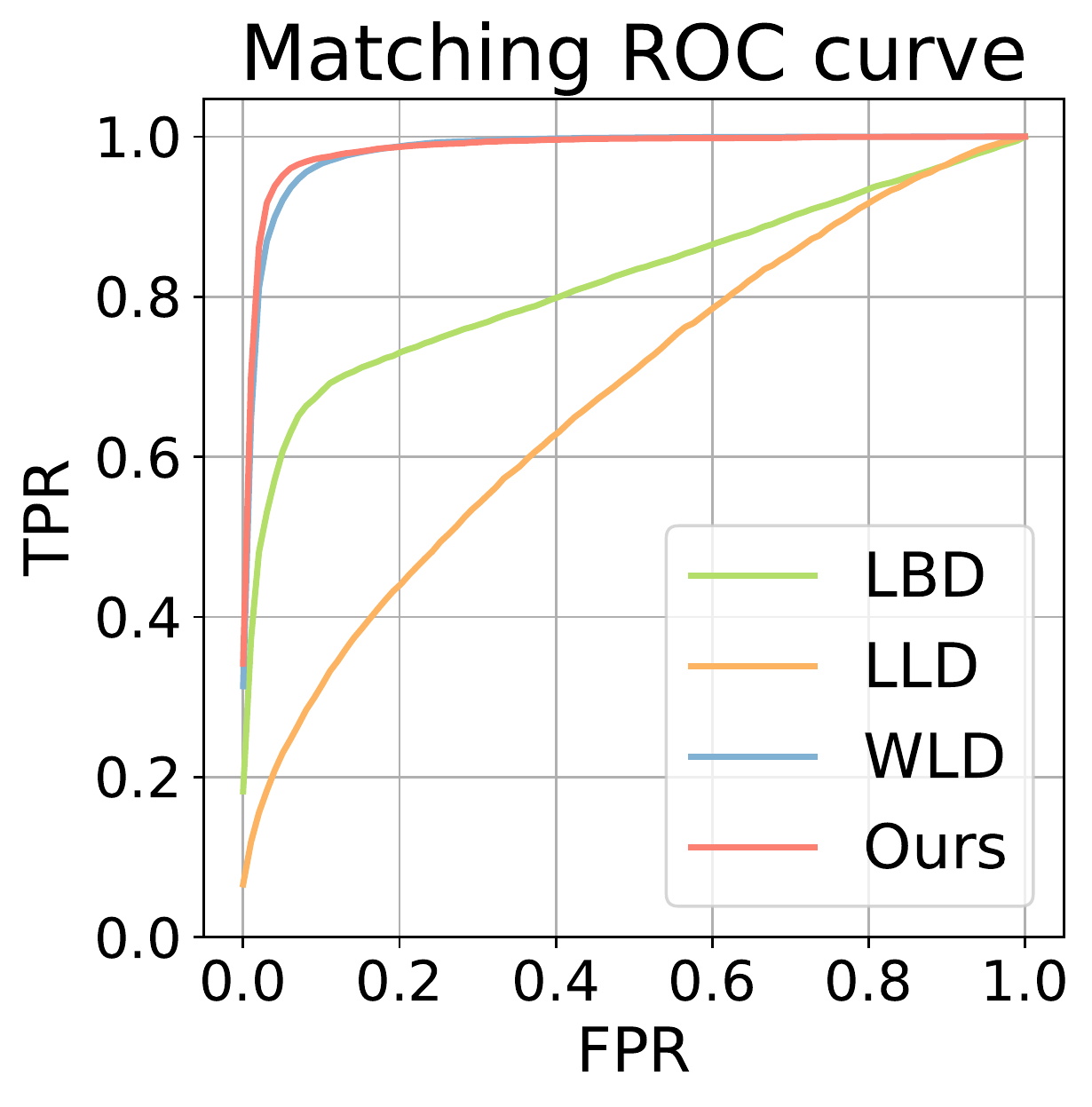}
        \vspace{-23pt}
    \end{minipage}
    \caption{\textbf{Descriptor evaluation on the Wireframe~\cite{wireframe} dataset with ground truth lines.} Matching the exact same lines yields a very high score for WLD and our method.}
    \label{fig:wireframe_matching_gt_lines}
\end{figure}

\boldparagraph{Descriptor evaluation on ground truth lines.}
Our first experiment aims at evaluating our approach on a perfect set of lines with a one-to-one matching. We thus use the Wireframe test set with its ground truth lines.
We compare our line matcher against 3 competing baselines: the handcrafted Line Band Descriptor (LBD)~\cite{zhang2013lbd}, the Learnable Line Descriptor (LLD)~\cite{vakhitov2019} and the Wavelet Line Descriptor (WLD)~\cite{lange2020wld}, an improved version of the Deep Line Descriptor (DLD)~\cite{lange2019dld}.
The results are shown in Figure~\ref{fig:wireframe_matching_gt_lines}.

Since LLD was trained on consecutive video frames with nearly no rotation between the images, it is not rotation invariant, hence its poor performance on the rotated images of our dataset. WLD showed that they were able to surpass the handcrafted LBD, and our descriptor gets a slight improvement over WLD by $5\%$.

\begin{figure}
    \begin{minipage}{0.7\columnwidth}
        \centering
        \includegraphics[width=\columnwidth]{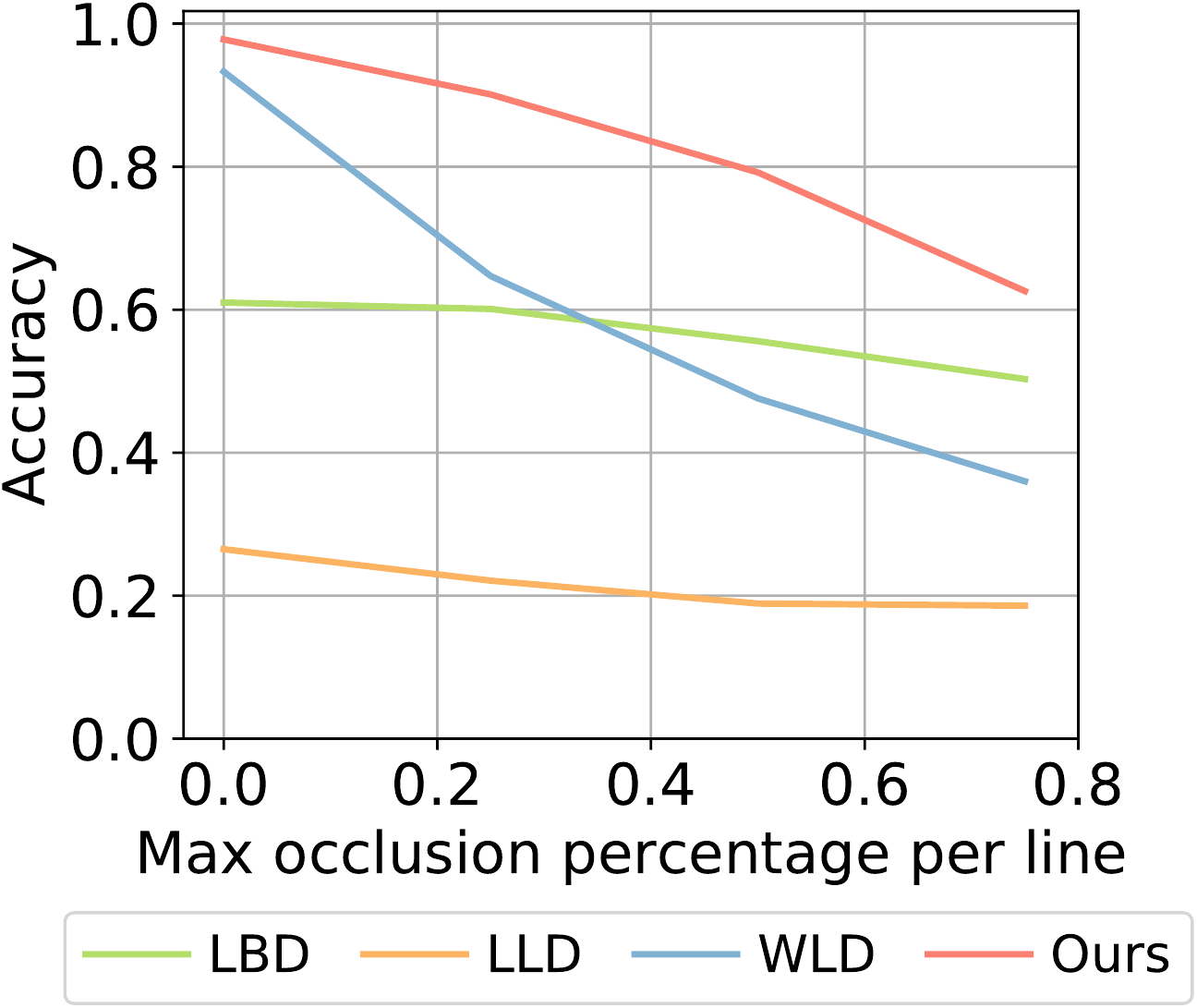}
        \vspace{-18pt}
    \end{minipage}
    \begin{minipage}{0.29\columnwidth}
        \centering
        \includegraphics[width=\columnwidth]{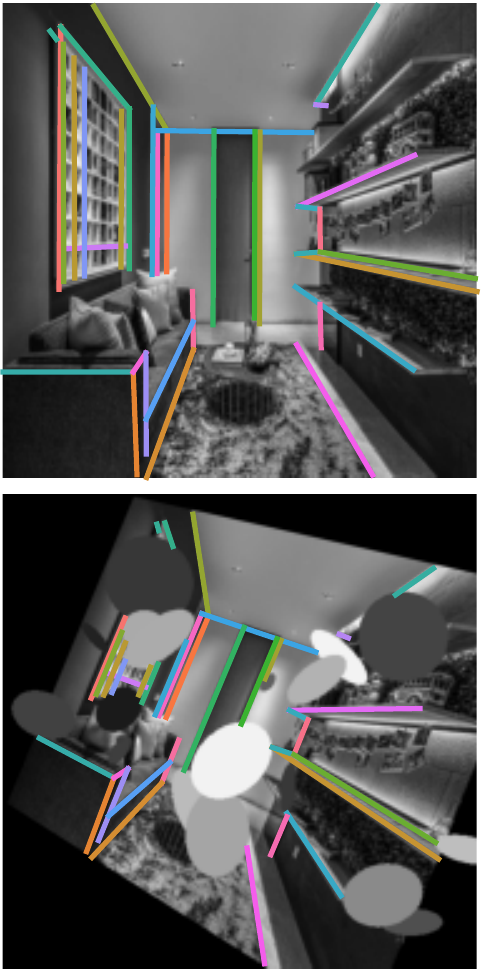}
        \vspace{-18pt}
    \end{minipage}
    \caption{\textbf{Robustness to occlusion.} \textbf{Left:} When evaluated on the Wireframe dataset with ground truth lines and random occluders, our method shows a higher robustness to occlusion compared to other methods. \textbf{Right:} Example of matches in the presence of occluders.}
    \label{fig:occlusion_exp}
\end{figure}

\boldparagraph{Robustness to occlusion experiment.}
In real-world applications, the detected lines across multiple views are rarely exactly the same, and some may be partially occluded or with different endpoints. To evaluate the robustness of our descriptor to these challenges, we modify the Wireframe test set to include artificial occluders.
More precisely, we overlay ellipses with random parameters and synthetic textures on the warped image of each pair, until at most $s\%$ of the lines are covered. We also shorten the line segments accordingly, so that each line stops at the occluders boundary. We compare line matches for various values of $s$ and get the results presented in Figure~\ref{fig:occlusion_exp}.
% More precisely, we perform two kinds of augmentation on the second image of each pair.
% First, half of the lines get shortened from one side by a factor $s$.
% Second, we overlay ellipses with random parameters and synthetic textures on the other half of the lines, covering at most $s\%$ of the line.
% Thus, in both cases the line segments are shortened or occluded by at most $s\%$.
% Figure~\ref{fig:occlusion_exp} displays the results of this experiment.

While all methods show a decrease in performance with a larger occlusion, \OURS{} outperforms the other baselines by a large margin for all degrees of occlusion. Note the significant drop for the learned baseline WLD, which operates on line patches and is thus severely affected by occlusions. This experiment thus validates the robustness of our method to occlusion and unstable line endpoints.

\begin{figure}
    \centering
    \includegraphics[width=\columnwidth]{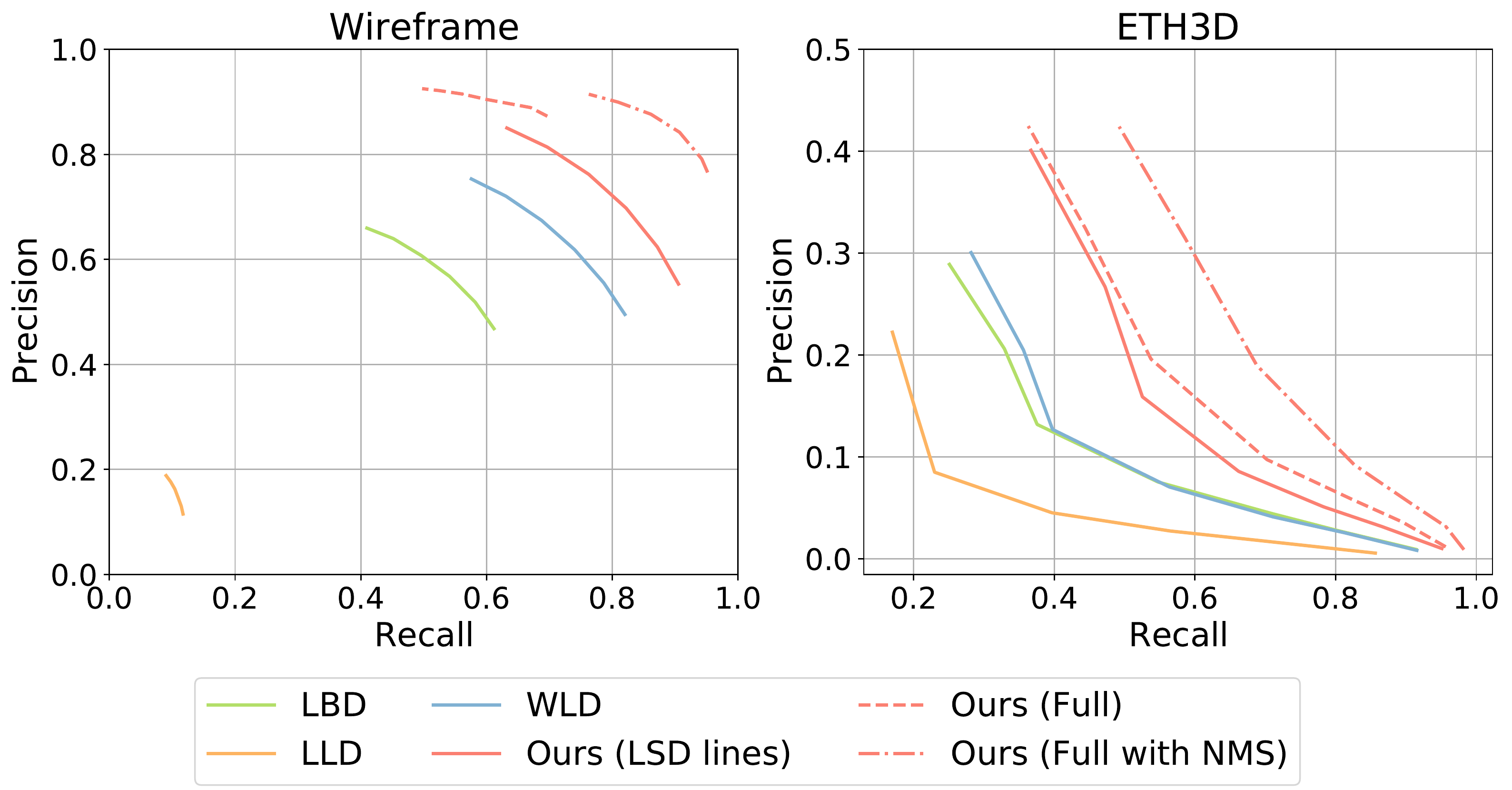}
    \vspace{-18pt}
    \caption{\textbf{Precision-Recall curves on predicted lines.} Our descriptor outperforms the other baselines when compared on LSD lines and the best performance is achieved for our full approach with our lines and descriptors.}
    \label{fig:pr_curves}
\end{figure}

\boldparagraph{Descriptor evaluation on predicted lines.}
To assess the performance of our proposed line description and matching, we also compute the matching metrics on predicted line segments instead of using hand-labeled lines. We perform two sets of experiments, on the Wireframe test set and on the ETH3D~\cite{eth3d} images which offer real world camera motions and can contain more challenging viewpoint changes than homographic warps. For the latter, images are downsampled by a factor of 8 and we select all pairs of images that share at least 500 covisible 3D points in the provided 3D models.
In both experiments, we run the LSD detector and compute all the line descriptor methods on them and also compare it with our full line prediction and description. Table~\ref{tab:descriptor_evaluation_pred_lines} and Figure~\ref{fig:pr_curves} evaluate the precision and recall of all methods.

Whether it is on synthetically warped images, or with real camera changes, \OURS{} outperforms all the descriptor baselines both in terms of matching precision and recall when compared on LSD lines. Using our own lines also improves the metrics, but the best performance is achieved when we apply a line NMS to remove overlapping segments. Having no overlap makes it indeed easier for the descriptor to discriminate the closest matching line.
\begin{table}[t]
    \centering
    %\footnotesize
    %\setlength{\tabcolsep}{2pt}
    \scriptsize
    \setlength{\tabcolsep}{4.2pt}
    \renewcommand{\arraystretch}{1}
    \begin{tabular}{llcccc}
        \toprule
        &                        & \multicolumn{2}{c}{Wireframe~\cite{wireframe}}        & \multicolumn{2}{c}{ETH3D~\cite{eth3d}} \\
        \cmidrule(lr){3-4} \cmidrule(lr){5-6}
        Lines & Desc                        & Precision$\uparrow$& Recall$\uparrow$ & Precision$\uparrow$ & Recall$\uparrow$ \\
        \midrule
        \multirow{4}{*}{LSD~\cite{von2008lsd}} & LBD~\cite{zhang2013lbd} & 0.496             & 0.597             & 0.132                                     & 0.376  \\
        & LLD~\cite{vakhitov2019}           & 0.123             & 0.116             & 0.085             & 0.230 \\
        & WLD~\cite{lange2020wld}           & 0.528             & 0.804             & 0.127             & 0.398 \\
        & \OURS{} (Ours)                    & \textbf{0.591}    & \textbf{0.889}    & \textbf{0.159}    & \textbf{0.525} \\[0.8pt] \hdashline \noalign{\vskip 1pt}
        Ours & \OURS{} (Ours)               & \textbf{0.882}    & 0.688             & \textbf{0.196}    & 0.538 \\
        Ours w/ NMS & \OURS{} (Ours)        & 0.777 & \textbf{0.949}    & 0.190             & \textbf{0.688} \\
        \bottomrule
    \end{tabular}
    \vspace{-4pt}
    \caption{\textbf{Matching precision and recall using LSD~\cite{von2008lsd} and our lines.} We use a threshold of 5 pixels in structural distance for the Wireframe~\cite{wireframe} images and of 5cm for the ETH3D images to define the ground truth matches.}
    \label{tab:descriptor_evaluation_pred_lines}
\end{table}

%------------------------------------------------------------------------------
\subsection{Ablation study}
%------------------------------------------------------------------------------
%
To validate the design choices of our approach, we perform an ablation study on the descriptor. \textit{SIFT endpoints} computes a SIFT descriptor~\cite{lowe2004sift} for both endpoints using the line direction as keypoint orientation, and averages the endpoints descriptor distance of each line candidate pair to get the line match scores. \textit{Average descriptor} computes a line descriptor by averaging the descriptors of all the points sampled along each line. \textit{NN average} computes the descriptor similarity of each line point with its nearest neighbor in the other line and averages all the similarities to get a line match score. \textit{D2-Net sampling} and \textit{ASLFeat sampling} refer to our proposed matching method where the points are sampled along the lines according to the saliency score introduced in D2-Net~\cite{d2net} and ASLFeat~\cite{aslfeat}, respectively. Finally, we test our method with various numbers of points sampled along each line. Table~\ref{tab:ablation_study} compares the accuracy of all these methods on the Wireframe dataset with ground truth lines both with and without occluders.

% \begin{table}[]
%     \centering
%     \footnotesize
%     \setlength{\tabcolsep}{10pt}
%     \renewcommand{\arraystretch}{1}
%     \begin{tabular}{lcc}
%         \toprule
%                             & \multicolumn{2}{c}{Matching accuracy$\uparrow$} \\ \cmidrule(lr){2-3}
%                             & GT lines          & GT lines w/ occluders \\
%         \midrule
%         SIFT~\cite{lowe2004sift} endpoints & 0.532             & 0.403 \\
%         Average descriptor                 & 0.944             & 0.754 \\
%         NN average                         & 0.972             & 0.803 \\
%         D2-Net~\cite{d2net} sampling       & 0.969             & 0.825 \\
%         ASLFeat~\cite{aslfeat} sampling    & 0.963             & 0.812 \\[0.8pt] \hdashline \noalign{\vskip 1pt}
%         Ours (3 samples)                   & \textbf{0.979}    & 0.813 \\
%         Ours (5 samples)                   & 0.978             & \textbf{0.846} \\
%         Ours (10 samples)                  & 0.972             & 0.836 \\
%         \bottomrule
%     \end{tabular}
%     \caption{\textbf{Ablation study on the Wireframe~\cite{wireframe} dataset.} We compare our method with various line matching and sampling methods along each line. Ground truth (GT) lines are used, both without occlusion and with up to $50\%$ occlusion.}
%     \label{tab:ablation_study}
% \end{table}
%
% side-by-side caption for tables, figures
% https://tex.stackexchange.com/questions/83897/putting-the-caption-of-a-table-on-its-right
\begin{table}[tb]
    \centering
    %\footnotesize
    %\setlength{\tabcolsep}{1pt}
    \scriptsize
    \setlength{\tabcolsep}{1.1pt}
    \renewcommand{\arraystretch}{1.02}
    \floatbox[\capbeside]{table}[0.53\columnwidth]%
    {\caption{\textbf{Ablation study on the Wireframe~\cite{wireframe} dataset.} We compare to various line matching and sampling methods along each line. Ground truth (GT) lines are used, both without occlusion and with up to $50\%$ occlusion.}
    \label{tab:ablation_study}}
    {\begin{tabular}{lcc}
        \toprule
                 & \multicolumn{2}{c}{Matching accuracy$\uparrow$} \\ \cmidrule(lr){2-3}
                 & GT lines          & GT lines \\
        Method   &                   & w/ occl. \\
        \midrule
        SIFT~\cite{lowe2004sift} endpoints & 0.532             & 0.403 \\
        Average descriptor                 & 0.944             & 0.754 \\
        NN average                         & 0.972             & 0.803 \\
        D2-Net~\cite{d2net} sampling       & 0.969             & 0.825 \\
        ASLFeat~\cite{aslfeat} sampling    & 0.963             & 0.812 \\[0.8pt] \hdashline \noalign{\vskip 1pt}
        Ours (3 samples)                   & \textbf{0.979}    & 0.813 \\
        Ours (5 samples)                   & 0.978             & \textbf{0.846} \\
        Ours (10 samples)                  & 0.972             & 0.836 \\
        \bottomrule
    \end{tabular}}
\end{table}

Results show that simply matching the line endpoints with a point descriptor such as SIFT is quickly limited and confirm the necessity of having a specific descriptor for lines. The small drop in matching accuracy for \textit{Average descriptor} and \textit{NN average} highlights the importance of keeping ordered points in NW matching. Surprisingly, smarter selections of points along each line such as \textit{D2-Net} and \textit{ASLFeat} sampling perform slightly worse than a regular sampling of points. Finally, there is a trade-off on the number of samples along each line: the NW algorithm loses its benefit when used with few points and the line descriptor becomes less robust to occlusions. On the other hand, many points along the line may produce descriptors that are too close from each other, which makes it harder to correctly discriminate between them. We found that 5 samples is a good trade-off overall, as was also the case for LLD~\cite{vakhitov2019}.

\begin{figure}[t]
    \centering
    \scriptsize
    \setlength{\tabcolsep}{0.3mm}
    \newcommand{\sz}{0.49}
    \begin{tabular}{cc}
        LSD~\cite{von2008lsd} + WLD~\cite{lange2020wld} & \OURS{} (Ours) \\
        \includegraphics[width=\sz\textwidth]{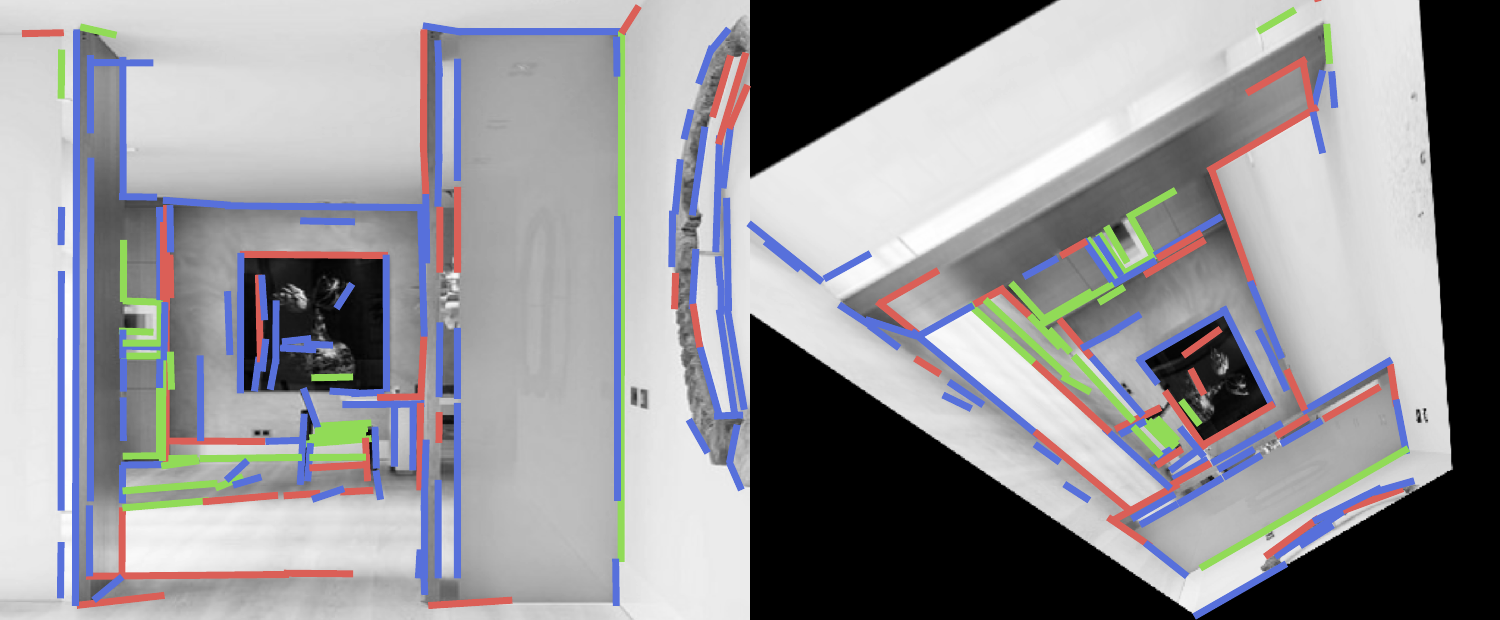} &
        \includegraphics[width=\sz\textwidth]{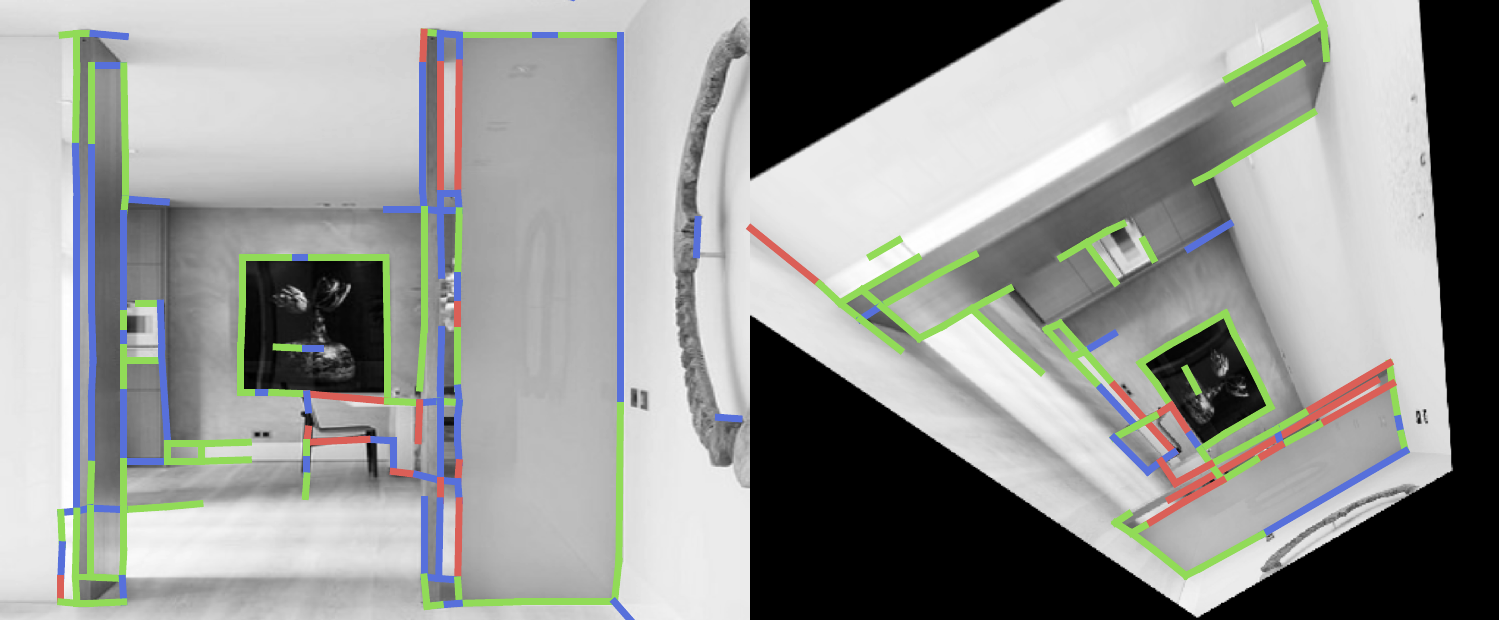}\\[-0.4mm]
        \includegraphics[width=\sz\textwidth]{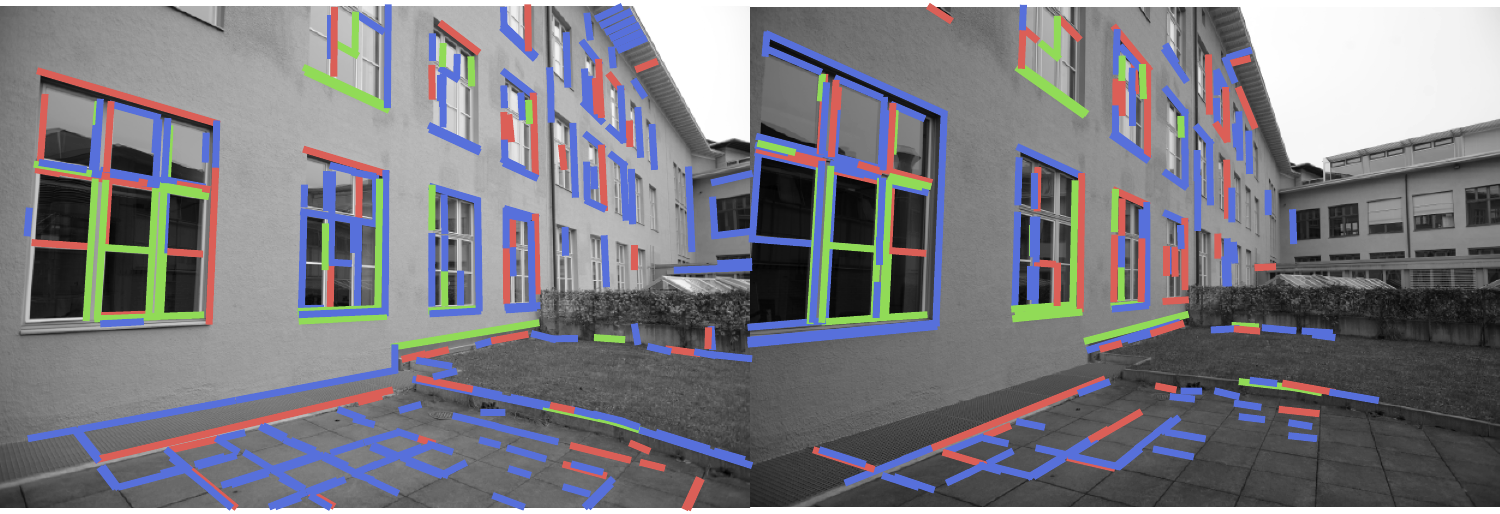} &
        \includegraphics[width=\sz\textwidth]{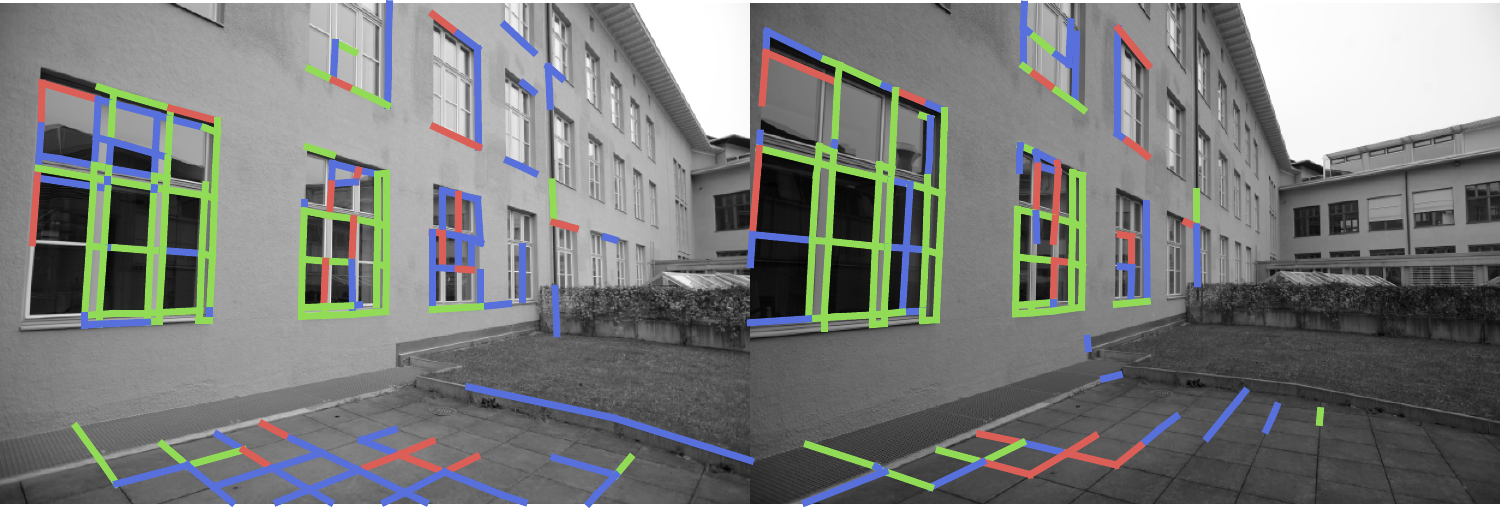}\\[-2mm]
    \end{tabular}
    \caption{\textbf{Line matches visualization.} Comparison of line matches between LSD~\cite{von2008lsd} + WLD~\cite{lange2020wld} and our method with \textcolor{green}{correct} matches, \textcolor{red}{incorrect} ones, and \textcolor{blue}{unmatched} lines. \OURS{} provides fewer but more repeatable lines that can be matched in poorly textured areas and with repetitive patterns.}
    \label{fig:qualitative}
\end{figure}

\section{Conclusion}
We presented the first deep learning pipeline for joint detection and description of line segments in images. 
Thanks to a self-supervised training scheme, our method can be applied to most image datasets, in contrast with the current learned line detectors limited to hand-labeled wireframe images.
Our descriptor and matching procedure addresses common issues in line description by handling partial occlusions and poorly localized line endpoints, while benefiting from the discriminative power of deep feature descriptors. 
By evaluating our method on a range of indoor and outdoor datasets, we demonstrate an improved repeatability, localization accuracy and matching performance compared to previous baselines.

While our line segment predictions are designed to be generic, further work is needed to tune them for specific applications. 
For instance, line-based localization may prefer short and stable lines, while 3D reconstruction and wireframe parsing may favor longer lines to get a better estimate of the dimensions of the scene. 
Thanks to our flexible line segment definition, a tuning of the line parameters allows to steer the output segments in one direction or another.
Overall, we hope that our full line detection and description pipeline is a first step to catch up with the more mature field of feature point matching, to be later able to combine both points and lines in a unified framework.

{\footnotesize
\vspace{0.5em}
\noindent \textbf{Acknowledgments.}
This work was supported by an ETH Zurich Postdoctoral Fellowship and Innosuisse funding (Grant No. 34475.1 IP-ICT).
}

\clearpage
\appendix
\appendixpage
In the following, we provide additional details about our Self-supervised Occlusion-aware Line Descriptor and Detector (\OURS{}). Section~\ref{appsec:synthetic_details} describes the generation of the synthetic dataset used to pretrain the network. Section~\ref{appsec:network_details} details our network architecture.
Section~\ref{appsec:multi_task} refers to the multi-task approach used to balance our different losses. Section~\ref{appsec:line_detection_details} explains into details some parts of the line segment detection module.
Section~\ref{appsec:evaluation_metrics} gives the exact equations used to compute the evaluation metrics considered in this work.
Section~\ref{appsec:statistics} provides proof that our results are statistically meaningful.
Section~\ref{appsec:eth3d} describes how we preprocessed the ETH3D dataset.
Section~\ref{appsec:homography_estimation} discusses the feasibility of applying our method to the homography estimation task.
% Section~\ref{appsec:3d_reconstruction} describes the accompanying demo video and the application of our line detection and matching to 3D reconstruction.
Finally section~\ref{appsec:qualitative_detection_matches} displays qualitative examples of our line detections and matches compared to other baselines.

\section{Synthetic dataset examples and homography generation}
\label{appsec:synthetic_details}

\begin{figure*}
\centering
    \includegraphics[width=\columnwidth]{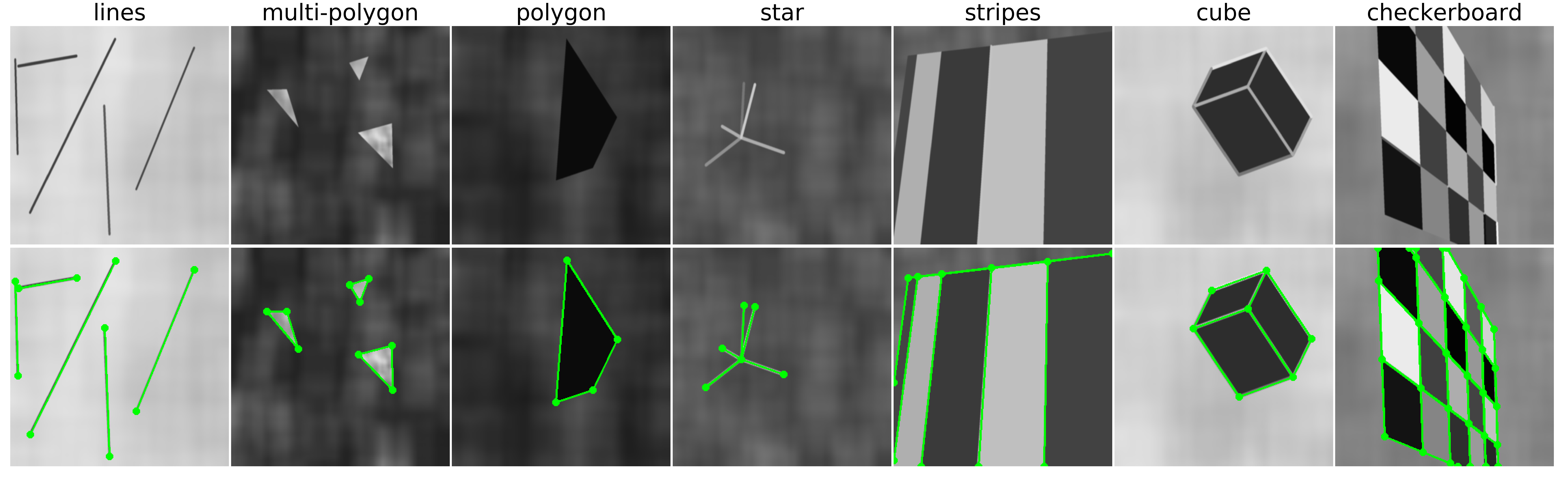}
    \caption{\textbf{Image examples from the synthetic dataset.} First row: rendered images. Second row: images with estimated junctions and line segment labels.}
    \label{appfig:synthetic_examples}
\end{figure*}

We provide here some examples of the images in our synthetic dataset and the homographies we used in both data augmentation and homography adaptation. These shapes include polygons, cubes, stars, lines, checkerboards, and stripes. Figure~\ref{appfig:synthetic_examples} shows some examples of these rendered shapes.

We follow the same process as in SuperPoint~\cite{superpoint} to generate the random homographies.
They are generated as a composition of simple transformations with pre-defined ranges: scaling (normal distribution $\mathcal{N}(1., 0.1)$), translation (uniform distribution within the image boundaries), rotation (uniformly in $[-90^{\circ}, +90^{\circ}]$), and perspective transform.
Examples of the difficulty of the test set can be observed in Figures~\ref{appfig:detection_qualitative} and \ref{appfig:comparison_superpoint} of the supplementary material.

\vspace{-2mm}

\section{Network architecture}
\label{appsec:network_details}

We provide here more details about our architecture and parameter choices. To have a fair comparison with most wireframe parsing methods~\cite{lcnn, hawp, deephough}, we use the same stacked hourglass network as in~\cite{hourglass}. Given an image with resolution $h \times w$, the output of the backbone encoder is a $\frac{h}{4} \times \frac{w}{4} \times 256$ feature map. The three heads of the network are implemented as follows:

\bolditem{Junction branch:} It is composed of a $3 \times 3$ convolution with stride 2 and 256 channels, followed by a $1 \times 1$ convolution with stride 1 and 65 channels, to finally get the $\frac{h}{8} \times \frac{w}{8} \times 65$ junction map.

\bolditem{Heatmap branch:} To keep a light network and avoid artifacts from transposed convolutions, we perform two consecutive subpixel shuffles~\cite{subpixel} blocks to perform a $\times4$ upsampling. More precisely, we use two $3 \times 3$ conv layers of output channel sizes 256 and 64, each of them followed by batch normalization, ReLU activation and a $\times 2$ subpixel shuffle for upsampling. A final $1 \times 1$ convolution with output channel 1 and sigmoid activation is then used to get the final line heatmap of resolution $h \times w$.

\bolditem{Descriptor branch:} The backbone encoding is processed by two consecutive convolutions of kernels $3 \times 3$ and $1 \times 1$, and output channels 256 and 128, to produce a $\frac{h}{4} \times \frac{w}{4} \times 128$ feature descriptor map. This semi-dense map can be later bilinearly interpolated at any point location. The triplet loss is optimized with a margin $M = 1$ and a minimal distance to the hardest negative of $T = 8$ pixels.

We use ReLU activations after each convolution and optimize the network with the Adam solver~\cite{kingma2014}. Images are resized to a $512 \times 512$ resolution and converted to grayscale during training.

\section{Multi-task learning}
\label{appsec:multi_task}
The tasks of detecting lines, their junctions, and describing them are diverse, and we assume them to have a different homoscedastic aleatoric uncertainty.
Additionally, they can have different orders of magnitude and their relative values are changing during training, in particular when the descriptor branch is added to the pre-trained detector network.
Therefore, we chose to use the multi-task loss introduced by Kendall \etal~\cite{uncertainty_dynamic} and successfully used in other geometrical tasks~\cite{Kendall2017,hfnet}, to automatically adjust the weights of the losses during training.

The final weights of Equation (8) gracefully converged towards the inverse of each loss, such that the value of each loss multiplied by its weight is around 1.
The final weight values are the following: $e^{-w_{junc}} = 7.2$, $e^{-w_{line}} = 16.3$ and $e^{-w_{desc}} = 8.2$. To show the effectiveness of the dynamic weighting, we tried two variants: (1) all loss weights are $1$, and (2) we used the final values from the dynamic weighting as static loss weights. In the first case, the detection and description results are worse by at least $10\%$ and $5.5\%$, respectively. In the second case, the detection and description results are worse by at least $6.7\%$ and $76.2\%$, respectively.

\section{Line segment detection details}
\label{appsec:line_detection_details}
To convert the raw junctions and line heatmap predicted by our network into line segments, the following steps are performed sequentially: regular sampling of points along each line, adaptive local-maximum search, and accepting the lines verifying a minimum average score and inlier ratio. Additionally, an initial step called candidate selection can be used to pre-filter some of the line candidates $\hat{L}$. We describe here two of these steps into more details: the adaptive local-maximum search, and the candidate selection.

\boldparagraph{Adaptive local-maximum search.}
Given a set of points sampled along a candidate line segment, one wants to extract the line heatmap values at these sampling locations. However, since the heatmap is limited to a resolution of one pixel, some samples may get a lower heatmap value if they land next to the actual line. Thus, we instead use an adaptive local-maximum search to find the highest activation of the line heatmap around each sampling location.
Given a line segment $\hat{l} = (\hat{e}^1, \hat{e}^2)$ from the candidate set $\hat{L}$ in an image of size $h \times w$, the search radius $r$ is defined as:
\begin{align}
    r = r_{min} + \lambda \frac{\norm{\hat{e}^1 - \hat{e}^2}}{\sqrt{h^2 + w^2}}
\end{align}
where $r_{min} = \frac{\sqrt{2}}{2}$ is the minimum search radius and $\lambda$ is a hyper-parameter to adjust the linear dependency on the segment lengths. We used $\lambda = 3$ pixels in all experiments.
The optimal line parameters were selected by a grid search on the validation set.
The $r_{min}$ parameter can in particular be kept constant across different image resolutions, without performance degradation.

\boldparagraph{Candidate selection (CS).}
In some application requiring line matching, having multiple overlapping segments may hinder the matching as the descriptor will have a harder job at discriminating close lines. Therefore, a non-maximum suppression (NMS) mechanism is necessary for lines. Unlike point or bounding box NMS, there is no well-established procedure for line NMS. Contrary to usual NMS methods which are used as postprocessing steps, we implement our line NMS as a preprocessing step, which actually speeds up the overall line segment detection as it removes some line candidates. Starting from the initial line candidates set $\hat{L}_{cand}$, we remove the line segments containing other junctions between their two endpoints. To identify whether a junction lies on a line segment, we first project the junction on the line and check if it falls within the segment boundaries. When it does, the junction is considered to be on the line segment if it is at a distance of less than $\xi_{cs}$ pixels from the line. Through out our experiments, we adopted $\xi_{cs} = 3$ pixels.

\section{Detector evaluation metrics}
\label{appsec:evaluation_metrics}
% Line segment detection metrics
% \noindent \textbf{Line segment detection metrics.}
Similarly to the metrics introduced in SuperPoint~\cite{superpoint}, we propose line segments repeatability and localization error metrics. Both of these metrics are computed using pairs of images $I_1$ and $I_2$, where $I_2$ is a warped version of $I_1$ under a homography $\mathcal{H}$. Each image is associated with a set of line segments $L_1=\{l_m^1\}_{m=1}^{M_1}$ and $L_2=\{l_m^2\}_{m=1}^{M_2}$, and $\mathrm{d}$ refers to one of the two line distances defined in this work: structural distance $\operatorname{d_s}$ and orthogonal distance $\operatorname{d_{orth}}$.

\bolditem{Repeatability:} The repeatability measures how often a line can be re-detected in different views. The repeatability with tolerance $\epsilon$ is defined as:
\begin{align}
    & \forall l \in L_1, \mathrm{C}_{L_2}(l) = 
    \begin{cases} 
    1\,\;\; \mathrm{if} (\min_{l_j^2 \in L_2} \mathrm{d}(l, l_j^2)) \leq \epsilon,\\
    0\,\;\; \text{otherwise}
    \end{cases} \\
    & \mathrm{Rep{\text -}\epsilon} = \frac{\sum_{i=1}^{M_1} \mathrm{C}_{L_2} (l_i^1) + \sum_{j=1}^{M_2} \mathrm{C}_{L_1}(l_j^2)}{M_1 + M_2}
\end{align}

\bolditem{Localization error:} The localization error with tolerance $\epsilon$ is the average line distance between a line and its re-detection in another view:
\begin{align}
    & \mathrm{LE\text{-}\epsilon} = \frac{\sum_{j\in Corr} \min_{l_i^1\in L1} \mathrm{d}(l_i^1, l_j^2)}{\abs{Corr}} \\
    & Corr = \{j \ |\ C_{L_1}(l_j^2) = 1 ,\ l_j^2 \in L_2\}
\end{align}
where $\abs{\cdot}$ measures the cardinality of a set.

\section{Statistical evaluation of our method}
\label{appsec:statistics}
\begin{figure}
    \begin{minipage}{0.6\columnwidth}
        \centering
        \tiny
        \setlength{\tabcolsep}{2.0pt}
        \renewcommand{\arraystretch}{1}
        \begin{tabular}{lcccc}
            \toprule
            & \multicolumn{2}{c}{$\operatorname{d_{s}}$}  & \multicolumn{2}{c}{$\operatorname{d_{orth}}$} \\
            \cmidrule(lr){2-3} \cmidrule(lr){4-5}
            & $\mathrm{Rep\text{-}5}$ $\uparrow$ & $\mathrm{LE\text{-}5}$ $\downarrow$
            & $\mathrm{Rep\text{-}5}$ $\uparrow$ & $\mathrm{LE\text{-}5}$ $\downarrow$ \\
            \midrule
            LCNN~\cite{lcnn}       & 0.336 & 2.777      & 0.637  & 1.878 \\
            HAWP~\cite{hawp}       & 0.451 & 2.625      & 0.537  & 1.738 \\
            DeepHough~\cite{deephough}  & 0.370 & 2.676      & 0.652  & 1.777 \\
            TP-LSD~\cite{huang2020tp}     & 0.563 & 2.467      & 0.746  & 1.450 \\
            LSD~\cite{von2008lsd}        & 0.358 & 2.079      & 0.707  & 0.825 \\
            \hdashline \noalign{\vskip 1pt}
            Ours (mean)                 & \textbf{0.577} & \textbf{1.955} & \textbf{0.891} & \textbf{0.804} \\
            Std deviation               & \textcolor{red}{0.020} & \textcolor{red}{0.070}
                                        & \textcolor{red}{0.0055} & \textcolor{red}{0.023} \\
            \bottomrule
        \end{tabular}
    \end{minipage}
    \begin{minipage}{0.37\columnwidth}
        \centering
        \includegraphics[width=0.8\textwidth]{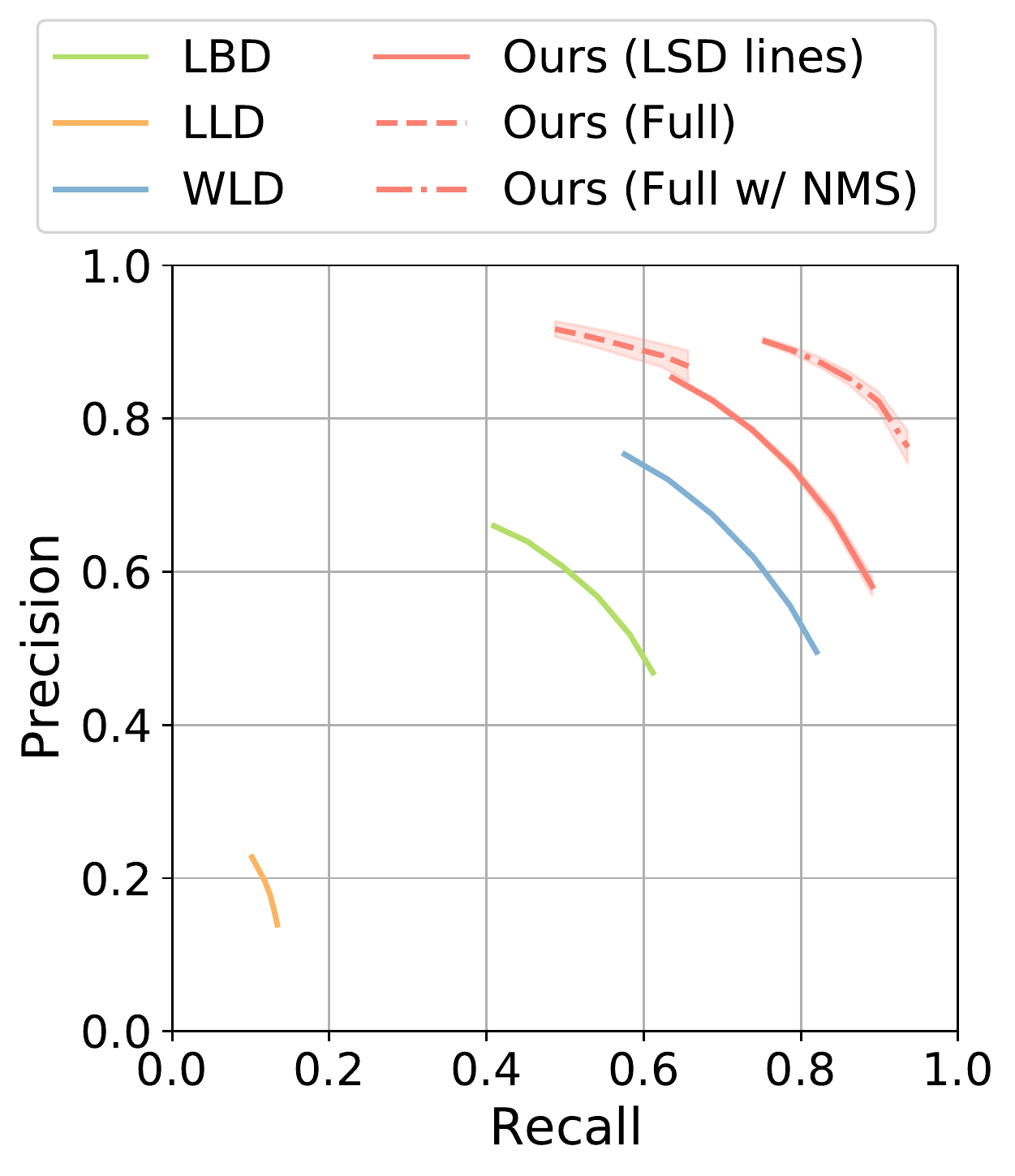}
    \end{minipage}
    \caption{\small{\textbf{Statistical significance of the evaluation.} We report the average value $\mu$ and standard deviation $\sigma$ of 5 different training runs of our method on the Wireframe dataset~\cite{wireframe}. Left: repeatability and localization error of the detector. Right: matching precision-recall curve with confidence interval $[\mu - \sigma, \mu + \sigma]$.}}
    \label{fig:statistical_evaluation}
\end{figure}

All the experiments displayed in the main paper are from a single training run.
To justify our statistical improvement over the baselines, we re-trained the full detector and descriptor network 5 times with different random seeds each time.
Figure~\ref{fig:statistical_evaluation} displays the same evaluation on the Wireframe dataset~\cite{wireframe} as in our paper with the mean and standard deviation of the performance of our method over these 5 runs, and thus shows statistically meaningful results.\\

\section{ETH3D dataset preprocessing}
\label{appsec:eth3d}
The ETH3D dataset~\cite{eth3d} is composed of 13 scenes taken in indoor as well as outdoor environments. Each image comes with the corresponding camera intrinsics and depth map, and a 3D model of each scene built with Colmap~\cite{schoenberger2016sfm} is provided as well. We use the undistorted image downsampled by a factor of 8 to run the line detection and description. We then use the depth maps and camera intrinsics to reproject the lines in 3D and compute the descriptor metrics in 3D space. While the depth maps have been obtained from a high-precision laser scanner, they contain some holes, in particular close to depth discontinuities. Since these discontinuities are actually where lines are often located, we inpaint the depth in all of the invalid areas at up to 10 pixels from a valid depth pixel. We used NLSPN~\cite{park2020non}, the current state of the art in deep depth inpainting guided with RGB images.

\begin{figure*}[t]
    \begingroup
    \centering
    \scriptsize
    \newcommand{\sz}{0.31}
    \setlength{\tabcolsep}{3pt}
    \begin{tabular}{cccc}
        \rotatebox{90}{\makecell{\hspace{12pt}LSD~\cite{von2008lsd}\\ \hspace{10pt}+ LBD~\cite{zhang2013lbd}}} & \includegraphics[width=\sz\textwidth]{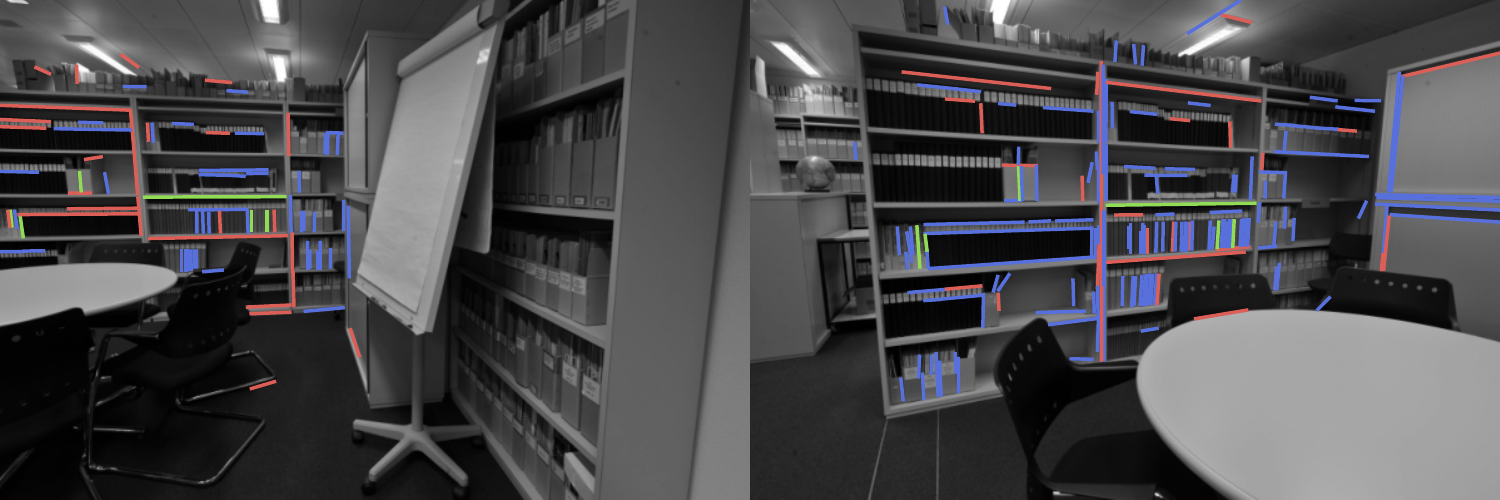} & \includegraphics[width=\sz\textwidth]{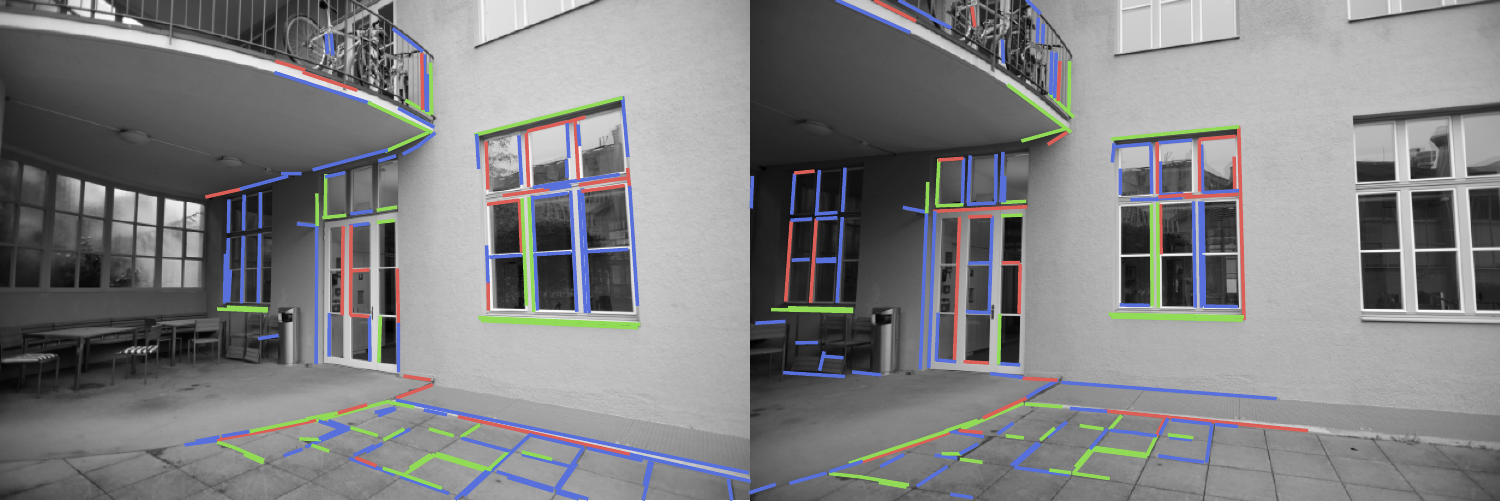} & \includegraphics[width=\sz\textwidth]{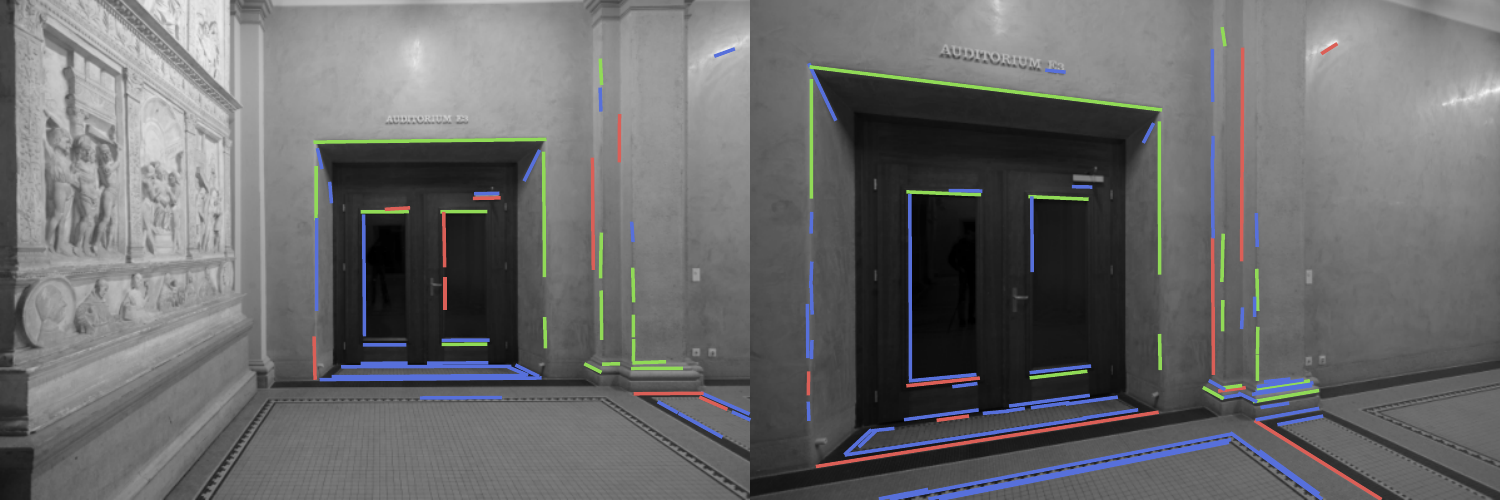}\\
        \rotatebox{90}{\makecell{\hspace{12pt}LSD~\cite{von2008lsd}\\ \hspace{10pt}+ LLD~\cite{vakhitov2019}}} & \includegraphics[width=\sz\textwidth]{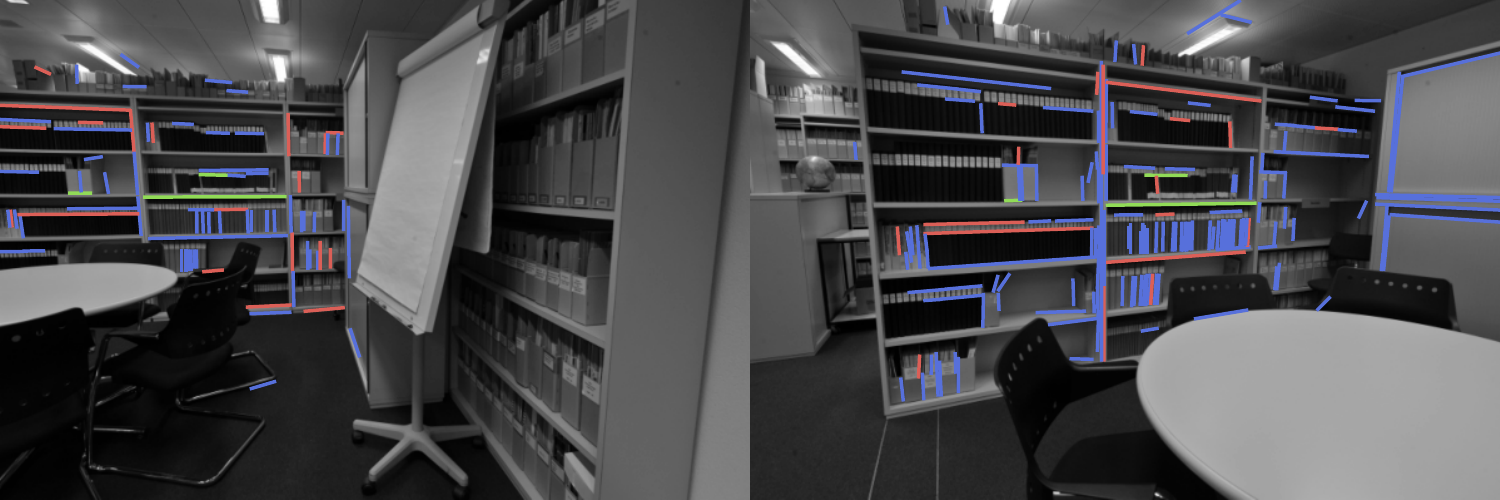} & \includegraphics[width=\sz\textwidth]{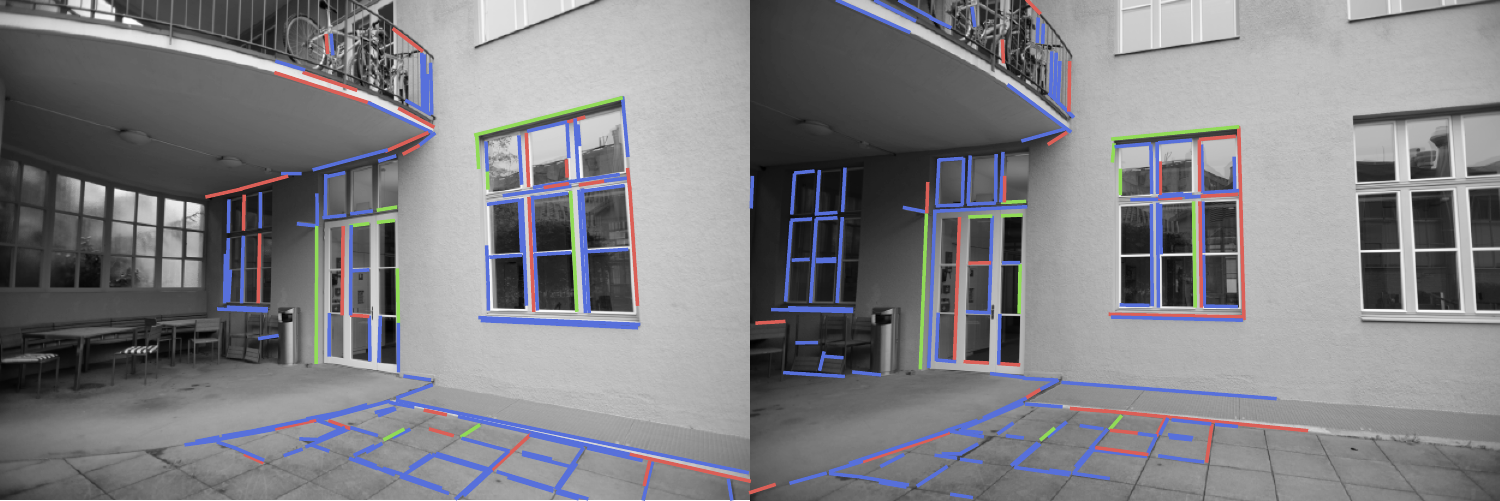} & \includegraphics[width=\sz\textwidth]{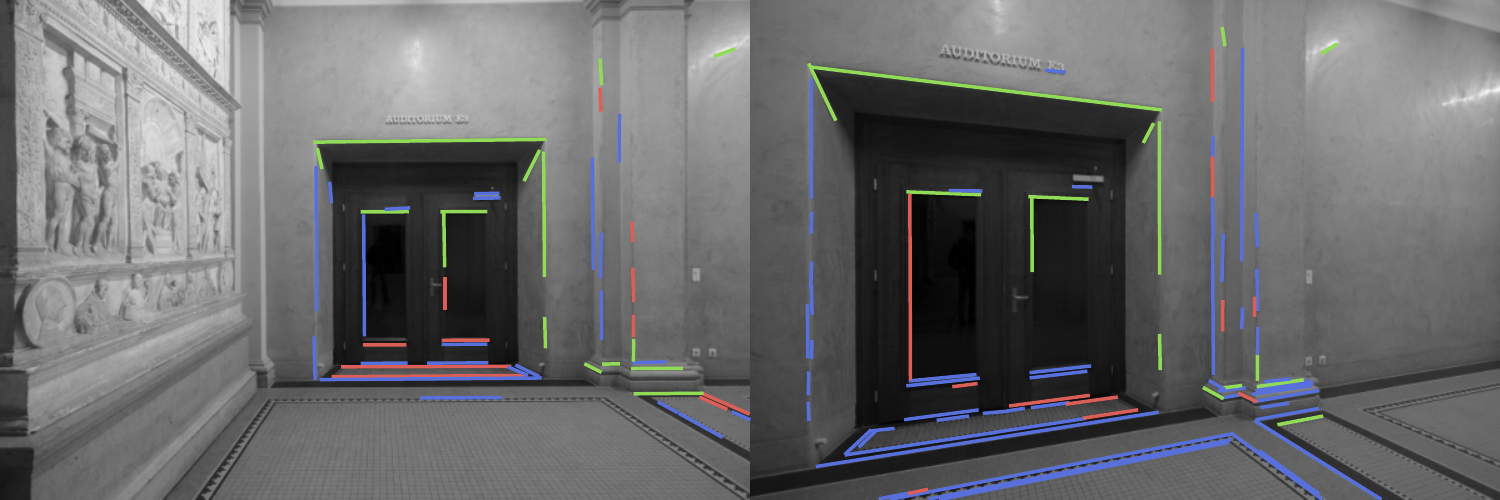}\\
        \rotatebox{90}{\makecell{\hspace{12pt}LSD~\cite{von2008lsd}\\ \hspace{10pt}+ WLD~\cite{lange2020wld}}} & \includegraphics[width=\sz\textwidth]{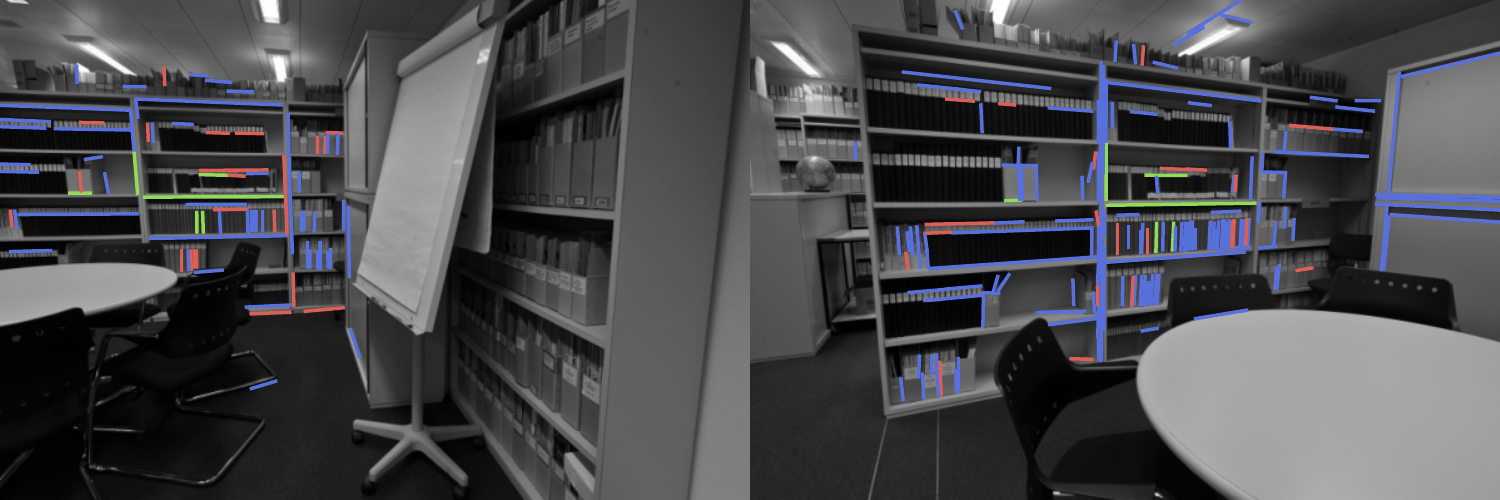} & \includegraphics[width=\sz\textwidth]{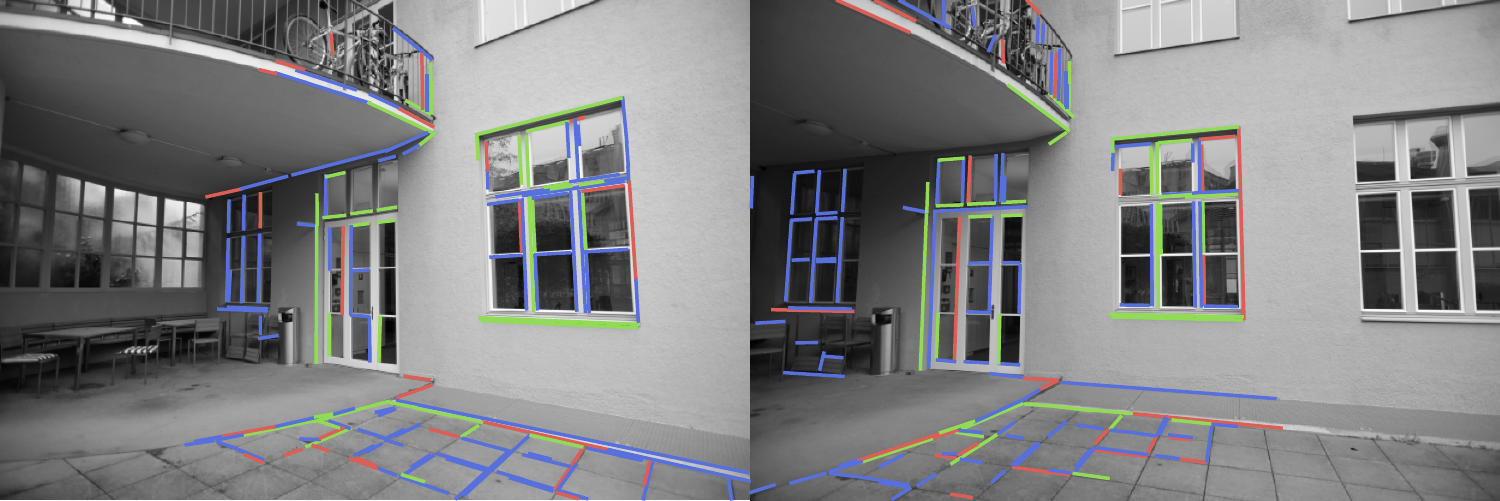} & \includegraphics[width=\sz\textwidth]{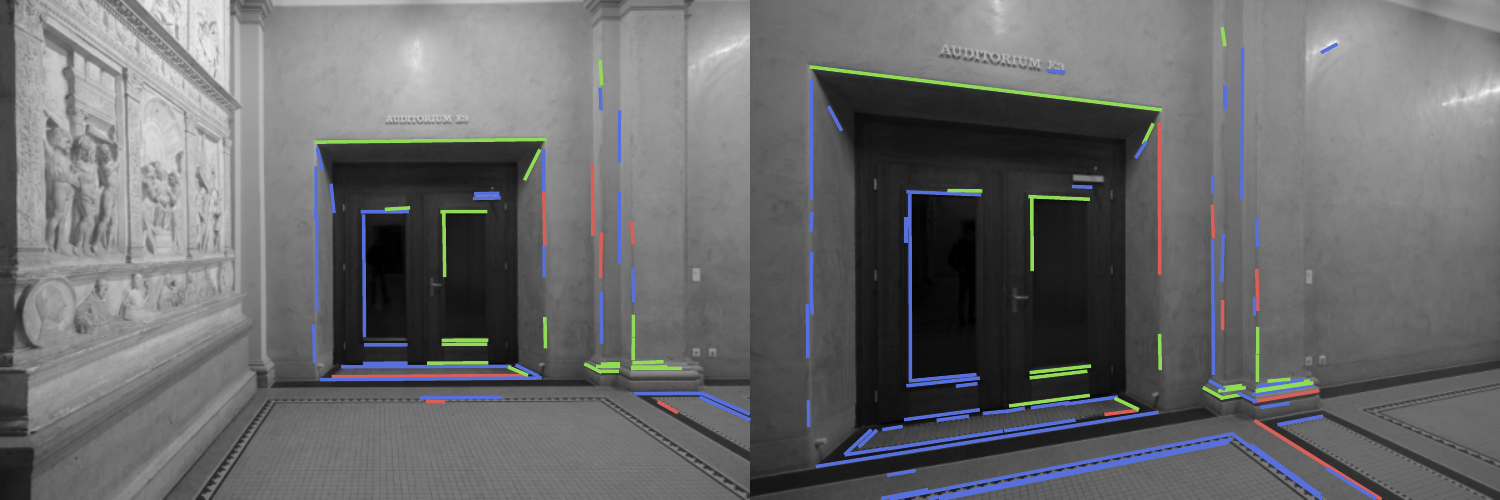}\\
        \rotatebox{90}{\hspace{16pt} Ours} & \includegraphics[width=\sz\textwidth]{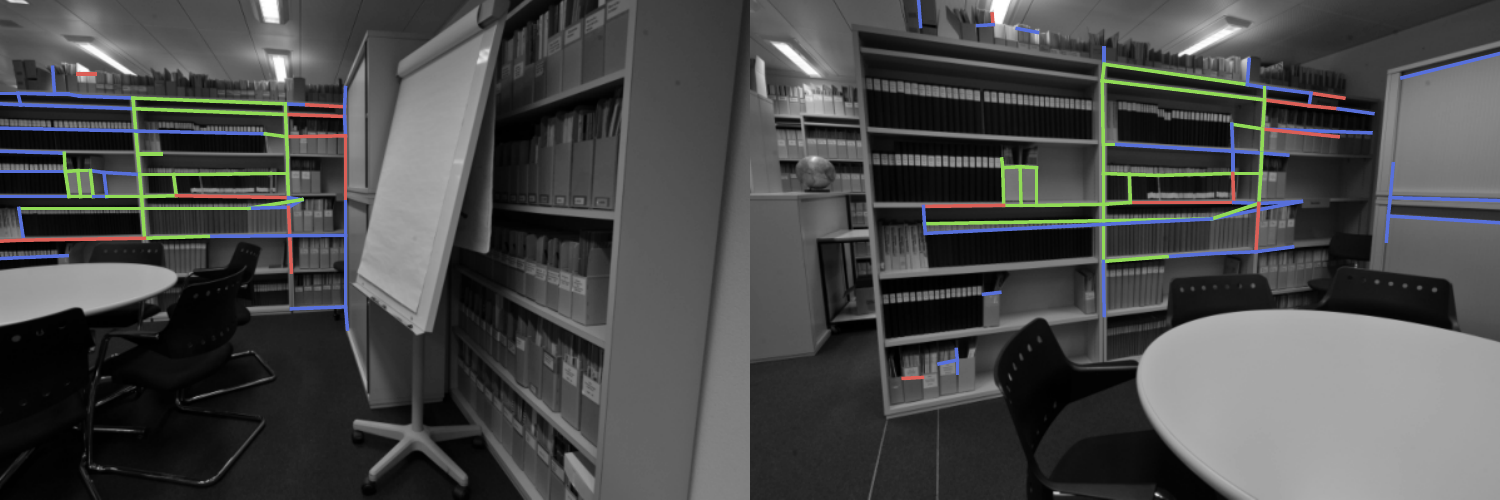} & \includegraphics[width=\sz\textwidth]{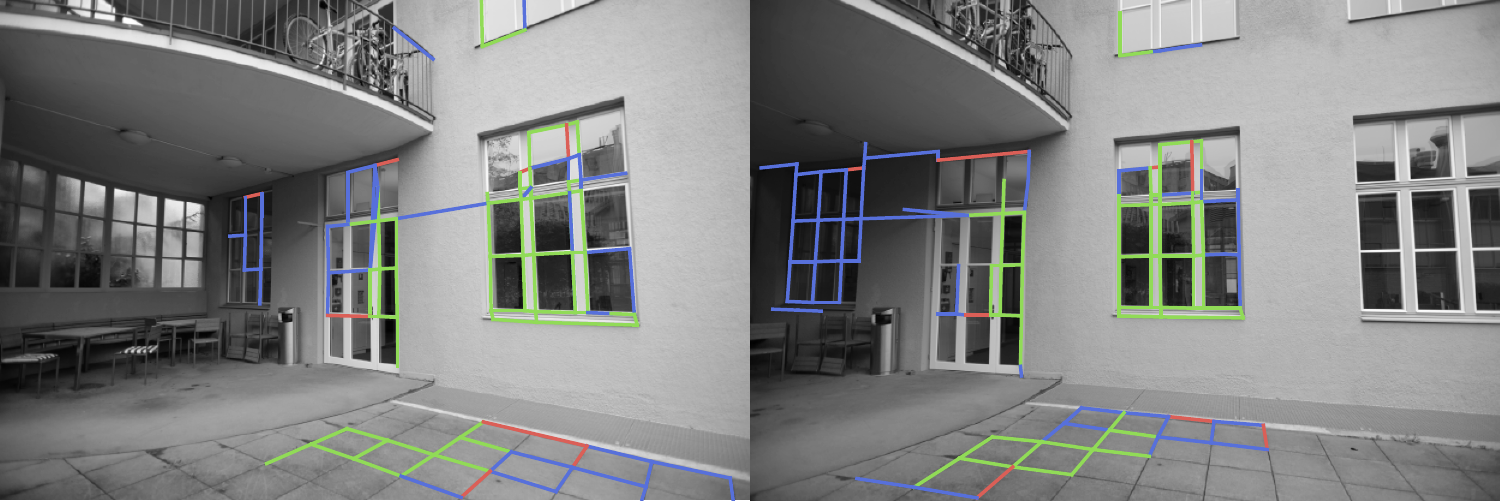} & \includegraphics[width=\sz\textwidth]{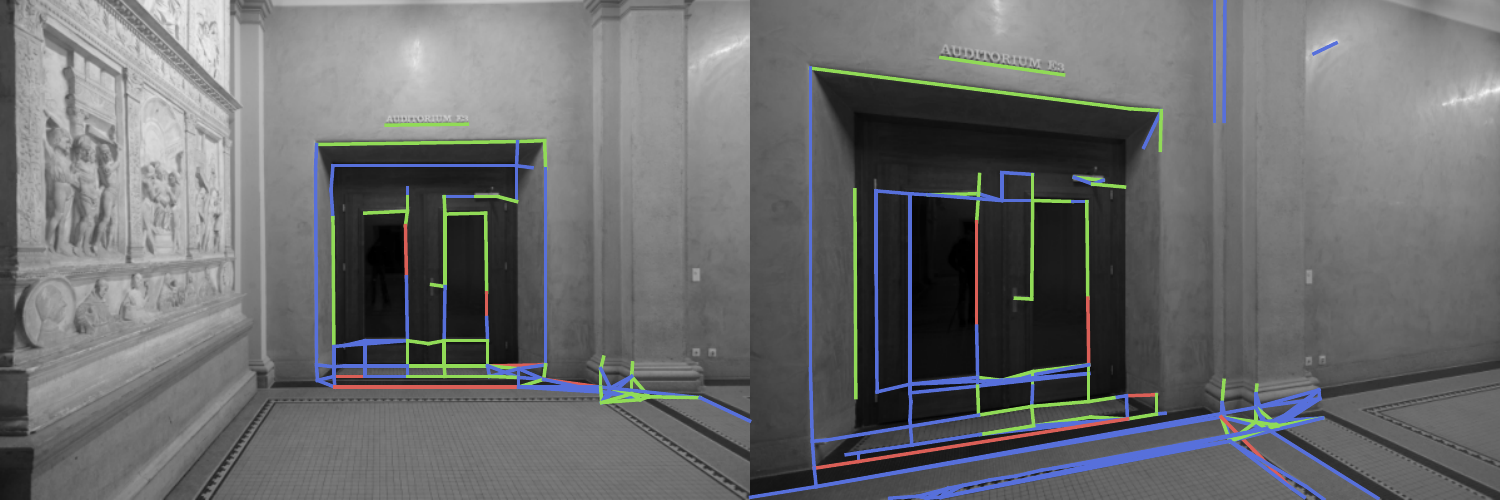}\\
    \end{tabular}
    \caption{\textbf{Qualitative results of line segment matching.} We display line segment matches on the ETH3D dataset~\cite{eth3d} with \textcolor{green}{correct matches}, \textcolor{red}{wrong matches} and \textcolor{blue}{unmatched lines}. Only lines shared between the two views are shown. Our full pipeline is compared to three line descriptor baselines computed on LSD lines~\cite{von2008lsd}: LBD~\cite{zhang2013lbd}, LLD~\cite{vakhitov2019} and WLD~\cite{lange2020wld}.} 
    \label{appfig:matches_qualitative}
    \endgroup
\end{figure*}

\begin{figure*}[t]
    \centering
    \scriptsize
    \setlength{\tabcolsep}{1.0pt}
    \renewcommand{\arraystretch}{0.8}
    \newcommand{\sz}{0.857}
    \begin{tabular}{cc}
        \rotatebox{90}{\hspace{+10mm}View2 \hspace{+18.5mm} View1} &
        \includegraphics[width=\sz\textwidth]{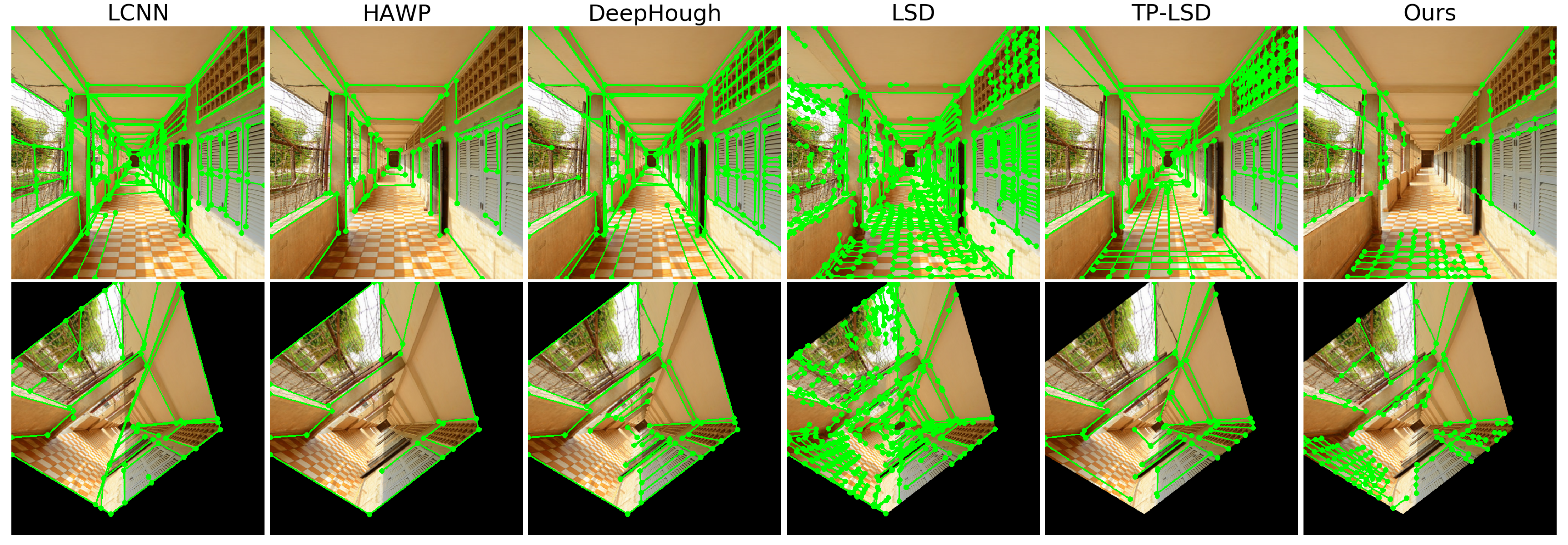}\\
        \midrule
        \rotatebox{90}{\hspace{+10mm}View2 \hspace{+18.5mm} View1} &
        \includegraphics[width=\sz\textwidth]{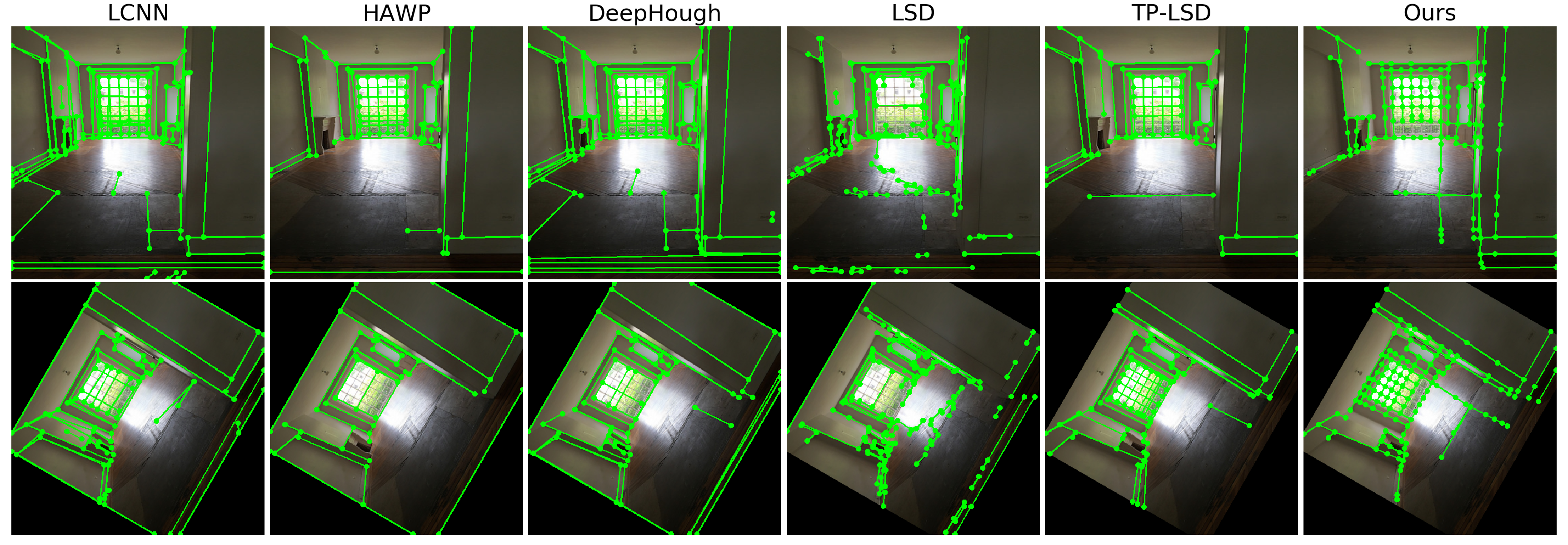}\\
        \midrule
        \rotatebox{90}{\hspace{+10mm}View2 \hspace{+18.5mm} View1} &
        \includegraphics[width=\sz\textwidth]{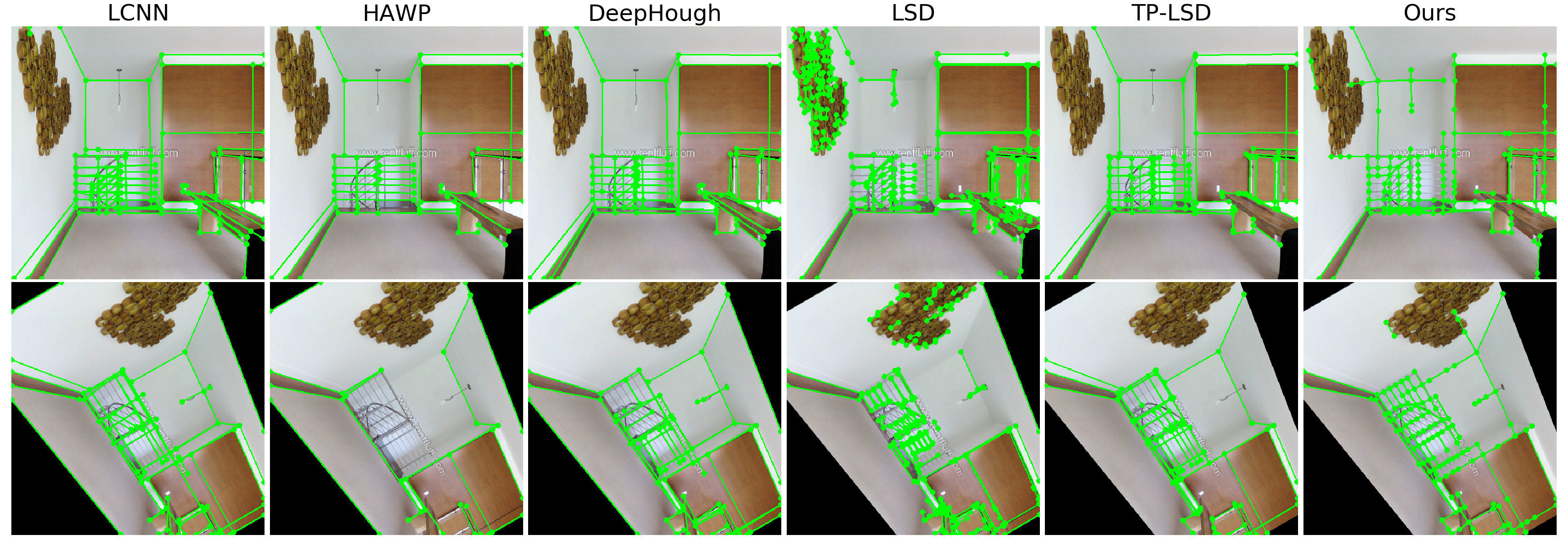}\\
        \midrule
        \rotatebox{90}{\hspace{+10mm}View2 \hspace{+18.5mm} View1} &
        \includegraphics[width=\sz\textwidth]{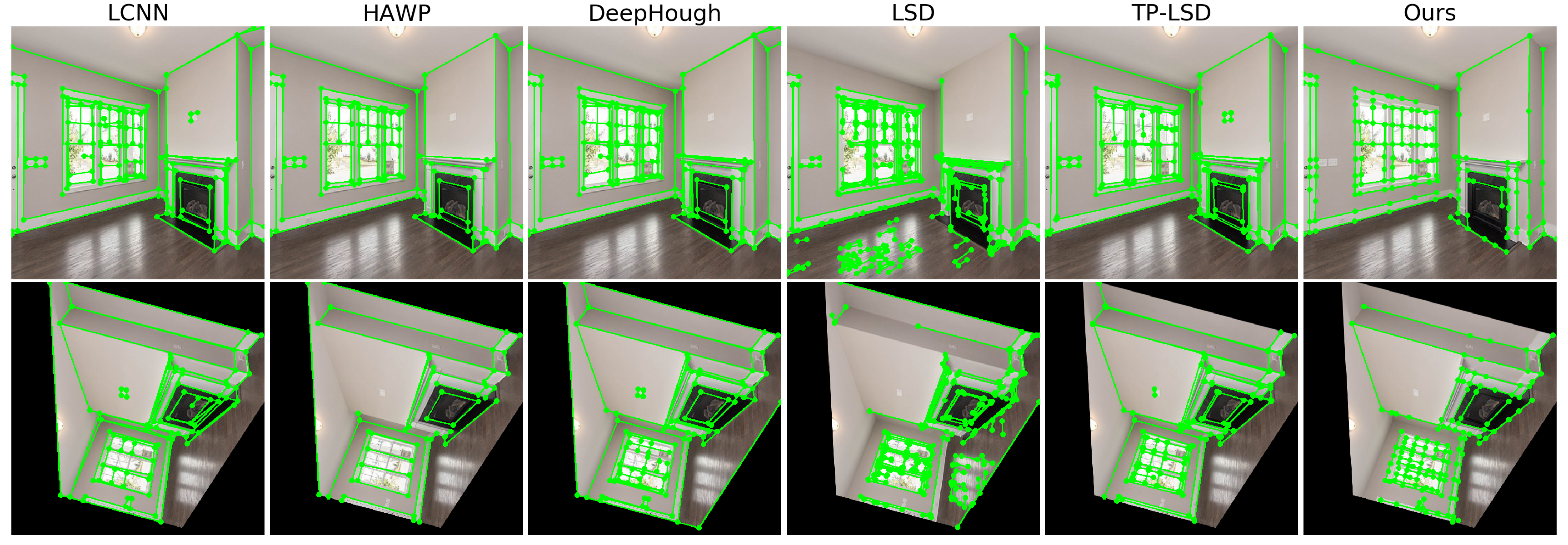}\\
    \end{tabular}
    \vspace{-9pt}
    \caption{\textbf{Qualitative results of line segment detections.} We show examples of line detections on the Wireframe dataset~\cite{wireframe} for the following methods: LCNN~\cite{lcnn}, HAWP~\cite{hawp}, DeepHough~\cite{deephough}, LSD~\cite{von2008lsd}, TP-LSD~\cite{huang2020tp} and ours.}
    \label{appfig:detection_qualitative}
\end{figure*}

\section{Homography estimation experiment}
\label{appsec:homography_estimation}

To validate the real-world applications of our method, we used the line segment detections and descriptors to match segments across pairs of images of the Wireframe dataset related by a homography~\cite{wireframe} and estimate the homography with RANSAC~\cite{fischler1981ransac}. We sample minimal sets of 4 lines to fit a homography and run up to $1,000,000$ iterations with the LORANSAC~\cite{Lebeda2012loransac} implementation of Sattler \etal\footnote{\url{https://github.com/tsattler/RansacLib}}. The reprojection error is computed with the orthogonal line distance. We compute the accuracy of the homography estimation similarly as in SuperPoint~\cite{superpoint} by warping the four corners of the image with the estimated homography, warping them back to the initial image with the ground truth homography and computing the reprojection error of the corners. We consider the estimated homography to be correct if the average reprojection error is less than 3 pixels. The results are listed in Table~\ref{apptab:homography_est}.

When compared on LSD line~\cite{von2008lsd}, our descriptors provide the highest accuracy among all baselines, and our full pipeline achieves a similar performance. When using our lines, we use a similar refinement of the junctions as in LSD~\cite{von2008lsd}: we sample small perturbations of the endpoints by a quarter of a pixel and keep the perturbed endpoints maximizing the line average score. Similarly to feature point methods~\cite{superpoint,d2net}, this experiment shows that learned features are still on par or slightly worse than handcrafted detections in terms of localization error.

We also add to the comparison the results of homography estimation for a learned feature point detector and descriptor, SuperPoint~\cite{superpoint}. The point-based approach performs significantly worse than our method, due to the numerous textureless scenes and repeated structures present in the Wireframe dataset. We also found that SuperPoint is not robust to rotations above 45 degrees, while our line descriptor can leverage its ordered sequence of descriptors to achieve invariance with respect to any rotation.

\begin{table}[t]
    \centering
    \scriptsize
    \setlength{\tabcolsep}{7.7pt}
    \renewcommand{\arraystretch}{1.1}
    \begin{tabular}{llccc}
        \toprule
              &                                             & \multicolumn{3}{c}{Homography estimation} \\ \cmidrule(lr){3-5}
        Lines & Desc                                        & Accuracy$\uparrow$  & \# inliers & Reproj. error$\downarrow$ \\
        \midrule
        \multirow{4}{*}{LSD~\cite{von2008lsd}}  & LBD~\cite{zhang2013lbd}   & 0.781             & 80      & 0.791 \\
                                                & LLD~\cite{vakhitov2019}   & 0.201             & 21      & 0.927 \\
                                                & WLD~\cite{lange2020wld}   & 0.920             & 116     & 0.868 \\
                                                & Ours                      & \textbf{0.948}    & 116     & 0.863 \\[0.8pt] \hdashline \noalign{\vskip 1pt}
        Ours                                    & Ours                      & 0.935             & 200     & \textbf{0.780} \\
        \midrule
        \multicolumn{2}{c}{SuperPoint~\cite{superpoint}}                    & 0.582             & 173      & 0.996 \\
        \bottomrule
    \end{tabular}
    \caption{\textbf{Evaluation results of homography estimation.} The homography between images of the Wireframe dataset~\cite{wireframe} is estimated from line matches using RANSAC. We use a threshold of 5 pixels in orthogonal line distance to consider a match to be an inlier.}
    \label{apptab:homography_est}
\end{table}

% \vspace{-2mm}
% \section{Demo video and 3D reconstruction application}
% \label{appsec:3d_reconstruction}
% \vspace{-2mm}
% We provide a demo video of our method, ``\textit{SOLD2.mp4}". In the video, we show (1) \OURS{} matches in the presence of occlusion, (2) \OURS{} matches on a hand-held device, and (3) an example of line-based 3D reconstruction using the segments from \OURS{}. For the line-based 3D reconstruction, we extended the incremental triangulator used in COLMAP~\cite{schonberger2016structure} to lines.

\section{Qualitative results of line segment detections and matches}
\label{appsec:qualitative_detection_matches}

\begin{figure*}[t]
    \begingroup
    \centering
    \scriptsize
    \newcommand{\sz}{0.45}
    \setlength{\tabcolsep}{3pt}
    \begin{tabular}{cc}
        SuperPoint~\cite{superpoint} & Ours\\
        \includegraphics[width=\sz\textwidth]{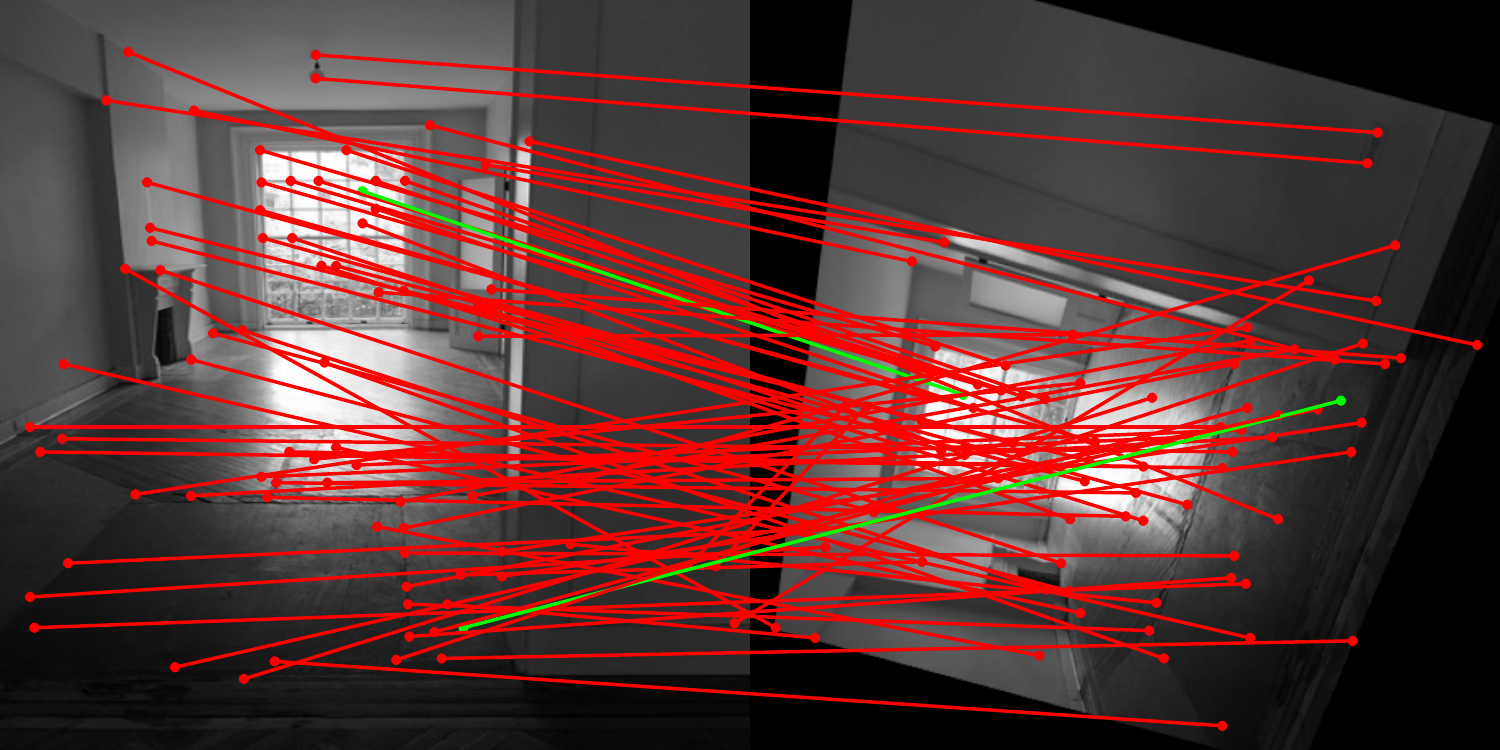} & \includegraphics[width=\sz\textwidth]{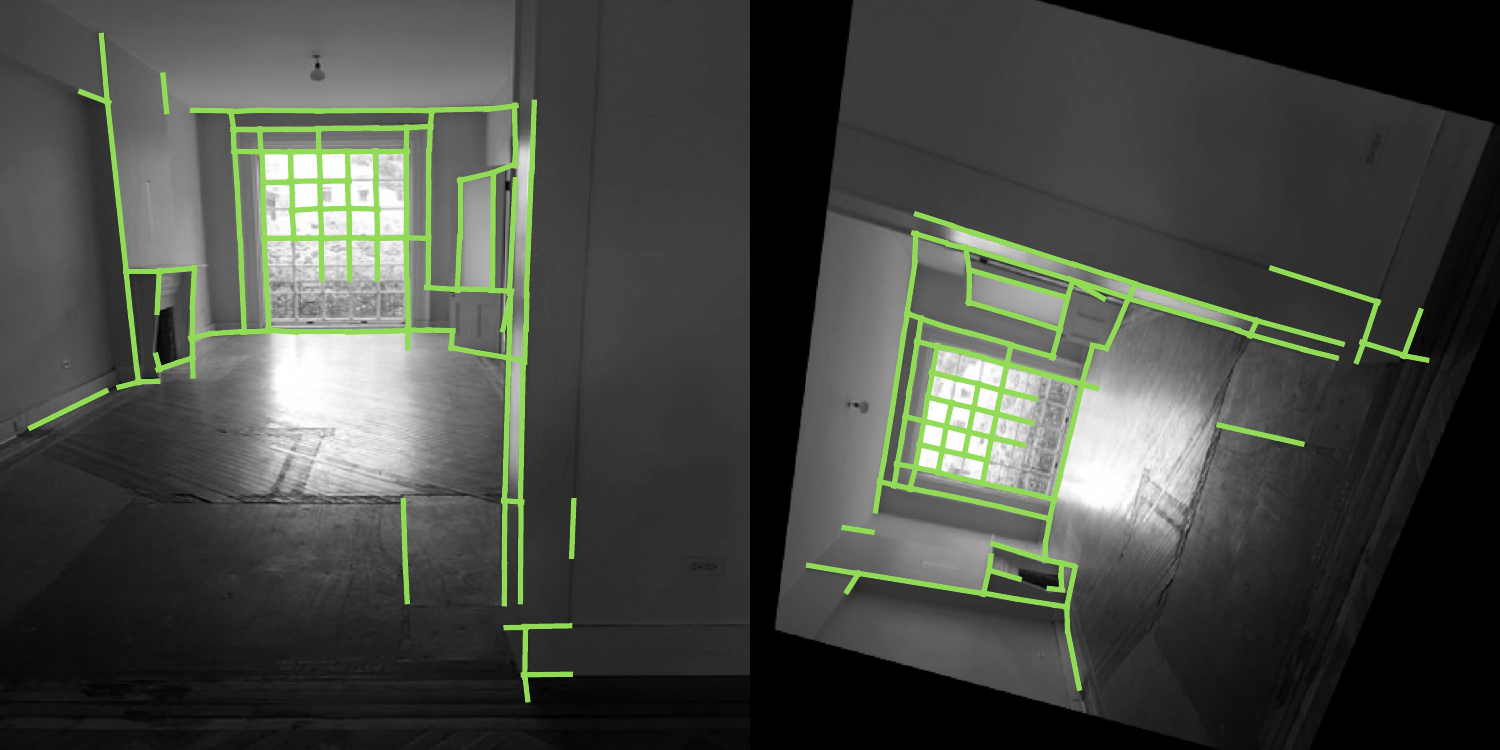}\\
        \includegraphics[width=\sz\textwidth]{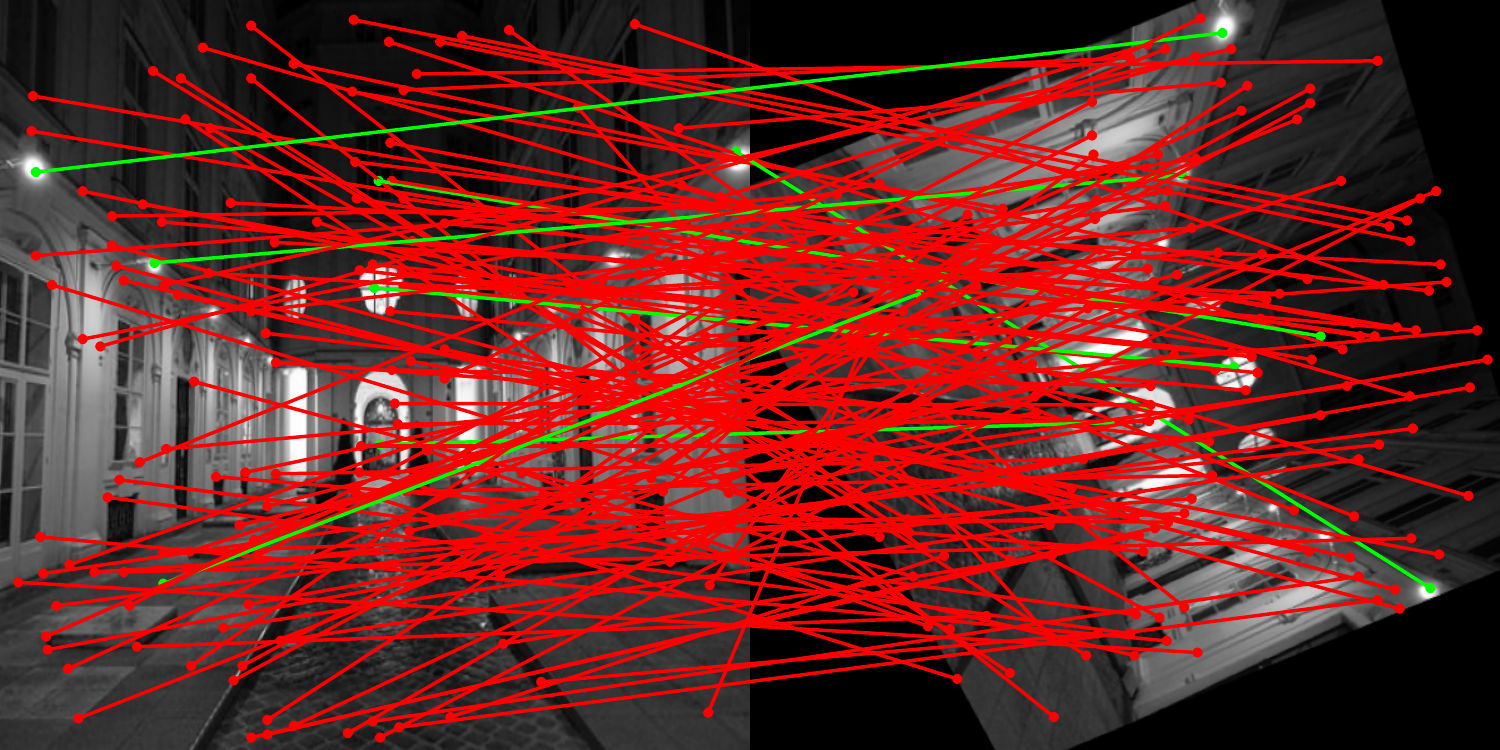} & \includegraphics[width=\sz\textwidth]{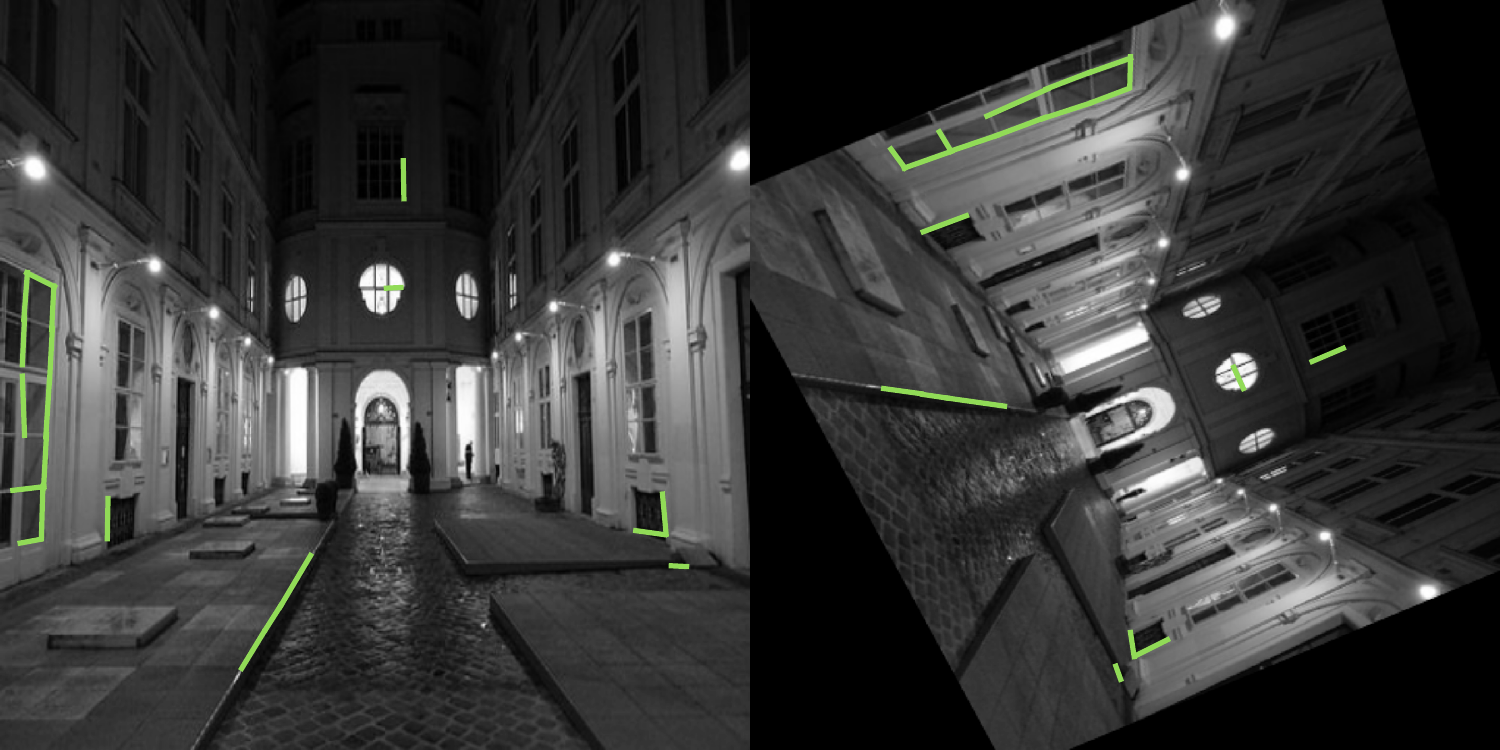}\\
        \includegraphics[width=\sz\textwidth]{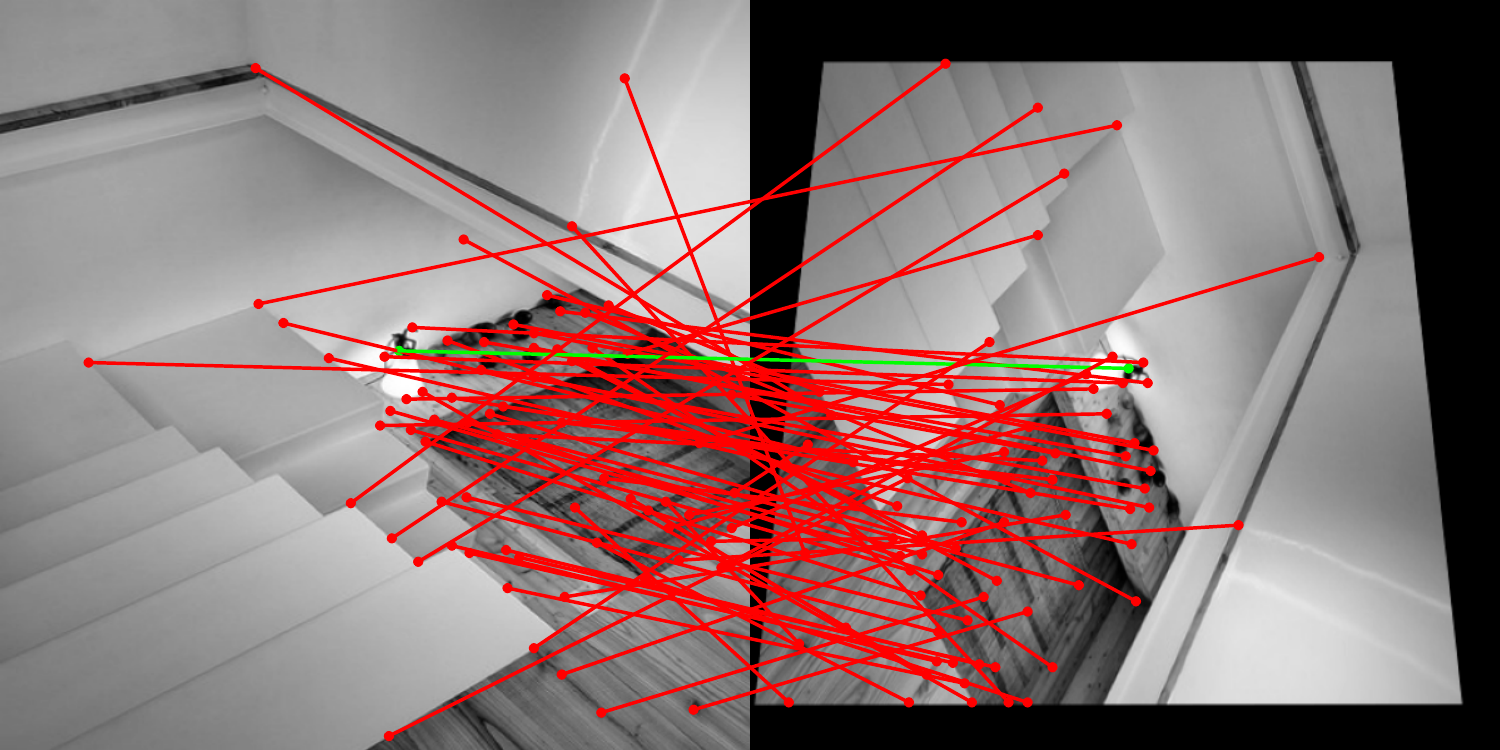} & \includegraphics[width=\sz\textwidth]{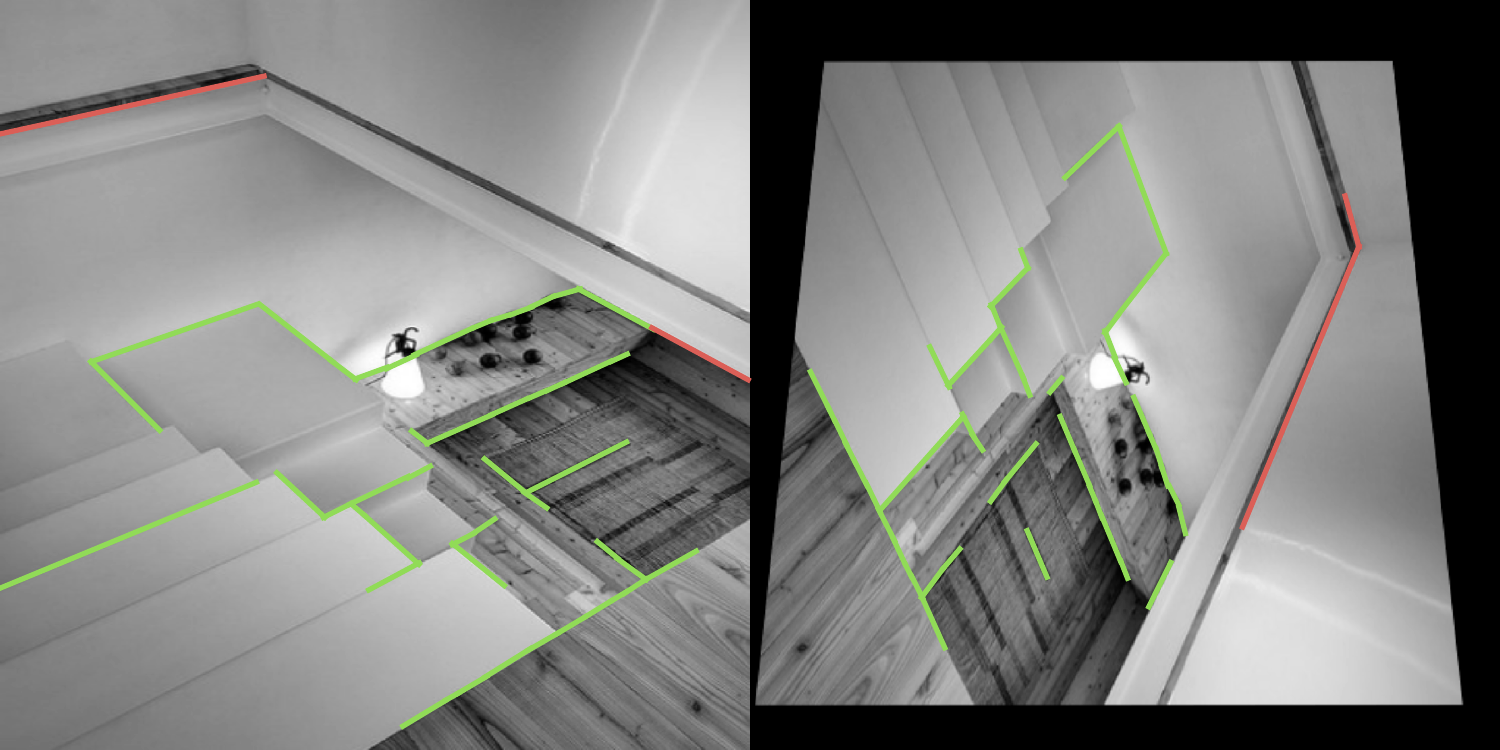}\\
        \includegraphics[width=\sz\textwidth]{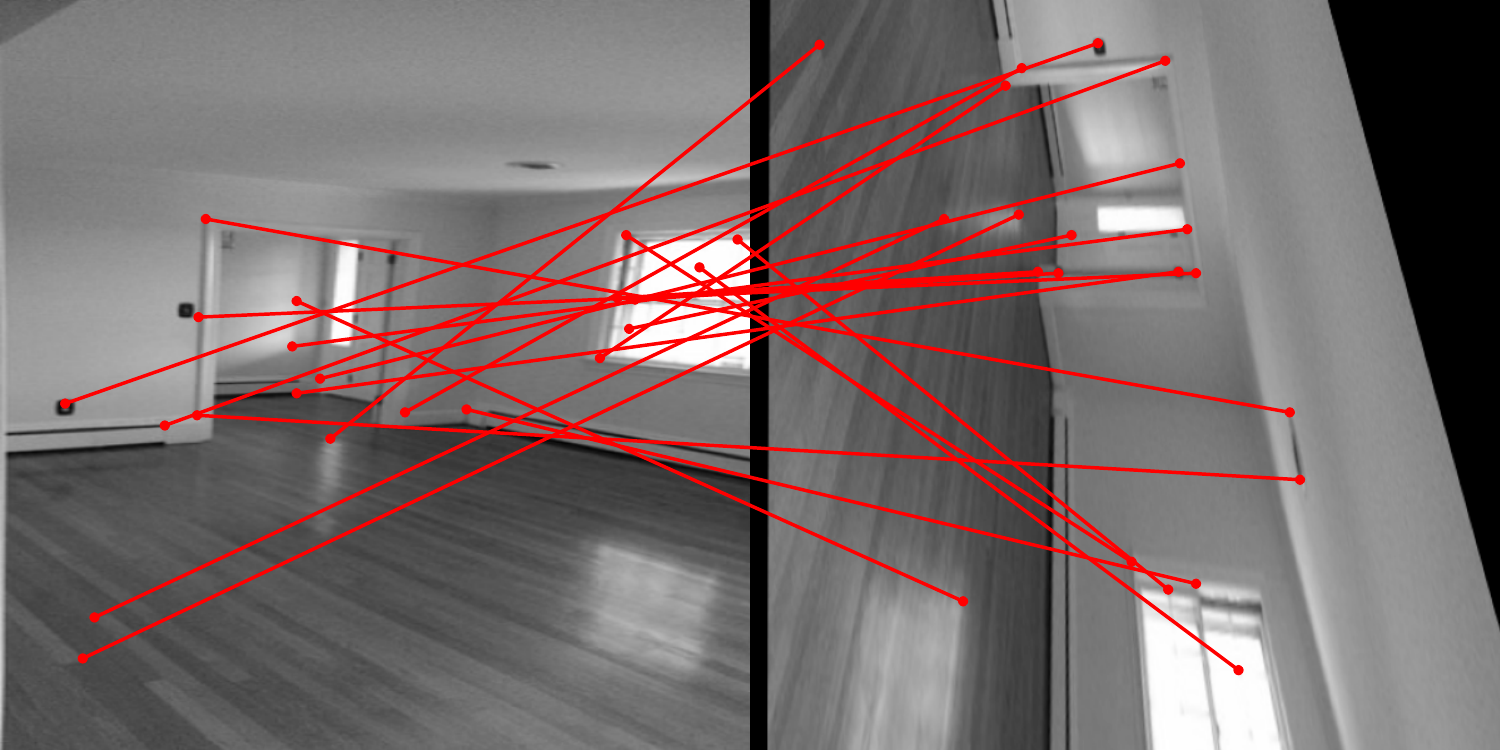} & \includegraphics[width=\sz\textwidth]{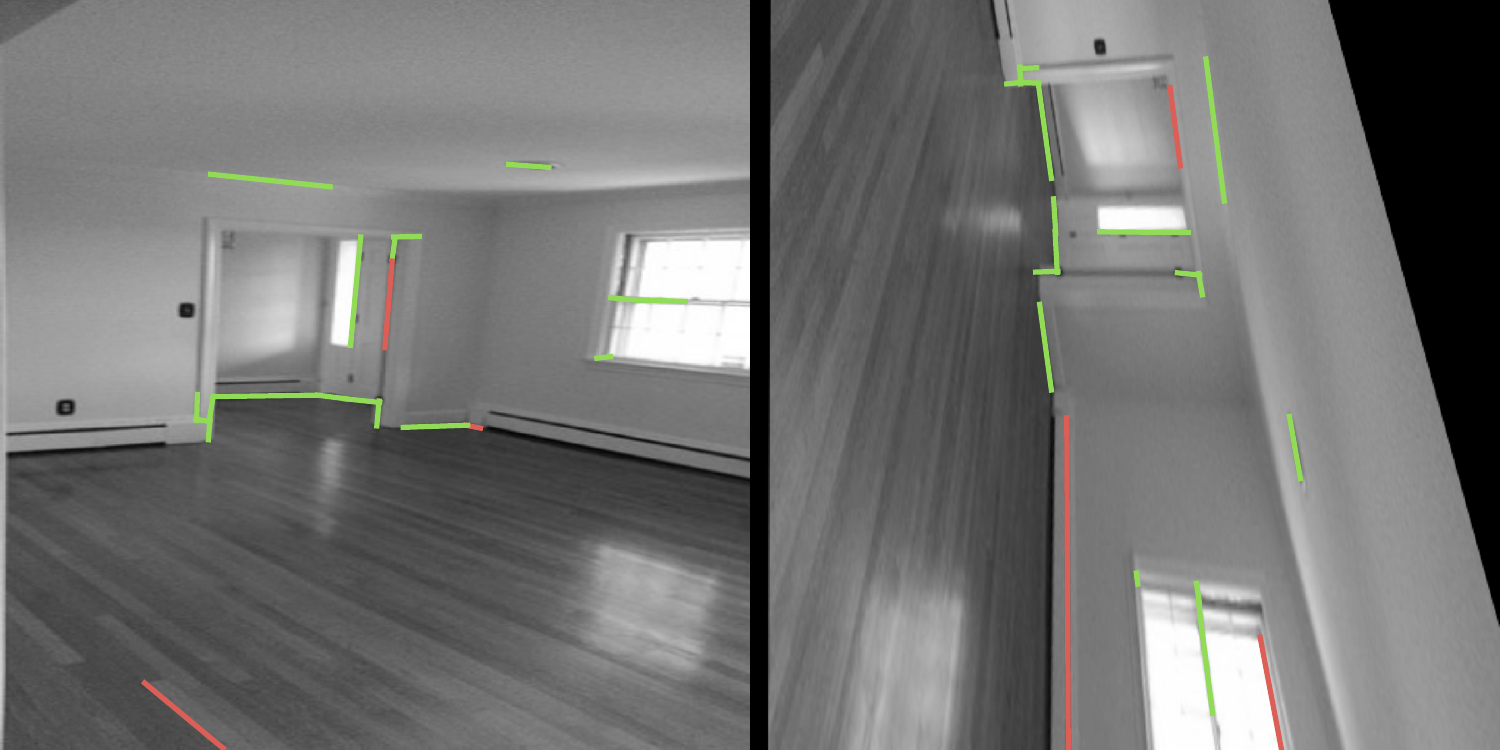}\\
        \includegraphics[width=\sz\textwidth]{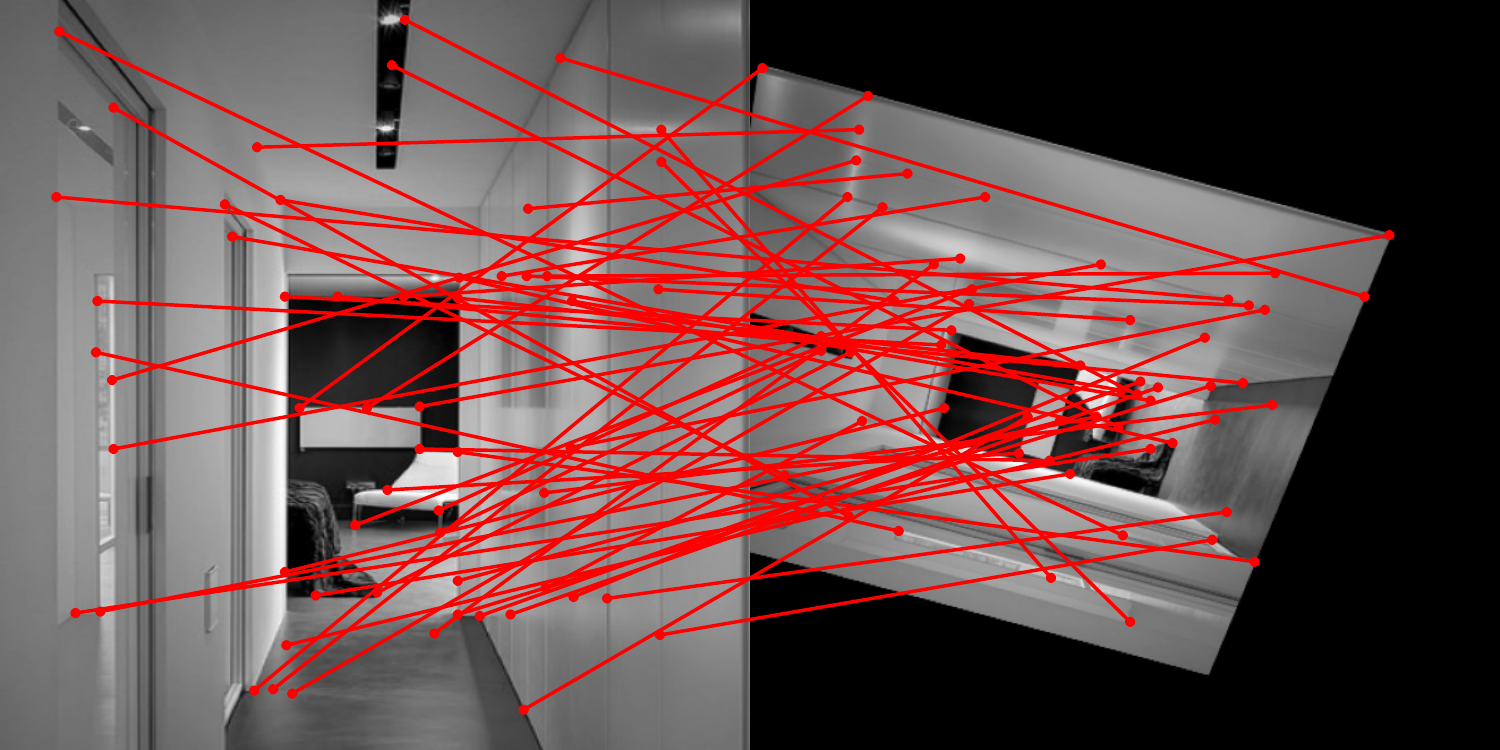} & \includegraphics[width=\sz\textwidth]{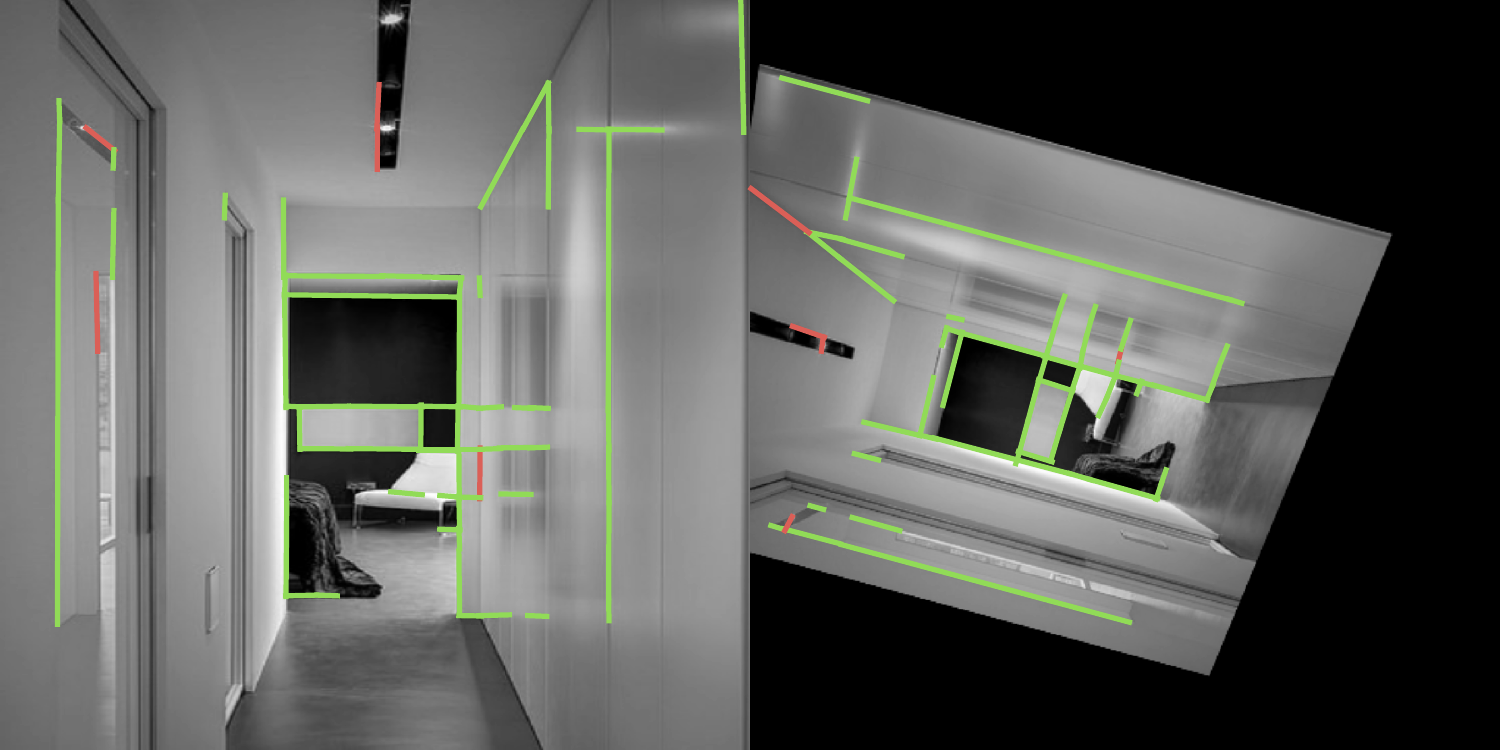}\\
    \end{tabular}
    \caption{\textbf{Benefits of lines compared to feature points.} We compare our method with point matching from SuperPoint~\cite{superpoint} on challenging images of the Wireframe dataset~\cite{wireframe} with \textcolor{green}{correct} and \textcolor{red}{wrong} matches. We use a distance threshold of 5 pixels to determine if a match is correct, using the orthogonal line distance in the case of lines. Lines can be matched even in the presence of textureless areas, as well as repeated and symmetrical structures.}
    \label{appfig:comparison_superpoint}
    \endgroup
\end{figure*}

We provide some visualizations of the line segment detection results in Figure~\ref{appfig:detection_qualitative} and of the line matching in Figure~\ref{appfig:matches_qualitative}. Figure~\ref{appfig:comparison_superpoint} also offers a comparison of line matches with point matches in challenging images with low texture, and repeated structures. Our method is able to match enough lines to obtain an accurate pose estimation, while point-based methods such as SuperPoint~\cite{superpoint} fail in such scenarios.

\clearpage
{\small
\bibliographystyle{ieee_fullname}
\bibliography{egbib}
}

\end{document}